\documentclass[conference]{IEEEtran}
\IEEEoverridecommandlockouts
\usepackage[nospace]{cite}
\usepackage{amsmath,amssymb,amsfonts}
\usepackage{graphicx}
\usepackage{textcomp}
\usepackage{xcolor}
\usepackage{subcaption}
\usepackage{bm}
\usepackage{url}
\usepackage{hyperref} 
\hypersetup{
    colorlinks=true,
    linkcolor=black,
    citecolor=black,
    urlcolor=black
}
\usepackage{multirow}
\usepackage{pifont}
\usepackage{algorithm}
\usepackage[noend]{algpseudocode}
\usepackage{amssymb}
\usepackage{amsthm}
\usepackage{footmisc}
\usepackage[flushleft]{threeparttable}

\usepackage{listings}
\usepackage{xcolor}
\definecolor{codegreen}{rgb}{0,0.6,0}
\definecolor{codegray}{rgb}{0.5,0.5,0.5}
\definecolor{codepurple}{rgb}{0.58,0,0.82}
\definecolor{backcolour}{rgb}{0.95,0.95,0.92}
\lstdefinestyle{mystyle}{
    backgroundcolor=\color{backcolour},   
    commentstyle=\color{codegreen},
    keywordstyle=\color{magenta},
    numberstyle=\tiny\color{codegray},
    stringstyle=\color{codepurple},
    basicstyle=\ttfamily\footnotesize,
    breakatwhitespace=false,         
    breaklines=true,                 
    captionpos=b,                    
    keepspaces=true,                 
    numbers=left,                    
    numbersep=5pt,                  
    showspaces=false,                
    showstringspaces=false,
    showtabs=false,                  
    tabsize=2
}
\newcommand\change[1]{\textcolor{black}{#1}}

\def\BibTeX{{\rm B\kern-.05em{\sc i\kern-.025em b}\kern-.08em
    T\kern-.1667em\lower.7ex\hbox{E}\kern-.125emX}}
\begin{document}

\title{DeAR: Accelerating Distributed Deep Learning with Fine-Grained All-Reduce Pipelining}

\author{
\IEEEauthorblockN{Lin Zhang\IEEEauthorrefmark{2}, Shaohuai Shi\IEEEauthorrefmark{3}\IEEEauthorrefmark{1}\thanks{*Corresponding author.}, Xiaowen Chu\IEEEauthorrefmark{4}\IEEEauthorrefmark{2}, Wei Wang\IEEEauthorrefmark{2}, Bo Li\IEEEauthorrefmark{2}, Chengjian Liu\IEEEauthorrefmark{5} \\}
\IEEEauthorblockA{\IEEEauthorrefmark{2}The Hong Kong University of Science and Technology, 
\IEEEauthorrefmark{3}Harbin Institute of Technology, Shenzhen,\\
\IEEEauthorrefmark{4}The Hong Kong University of Science and Technology (Guangzhou), 
\IEEEauthorrefmark{5}Shenzhen Technology University\\
lzhangbv@connect.ust.hk, shaohuais@hit.edu.cn, xwchu@ust.hk, \{weiwa, bli\}@cse.ust.hk, liuchengjian@sztu.edu.cn}
}
\maketitle
\begin{abstract}
Communication scheduling has been shown to be effective in accelerating distributed training, which enables all-reduce communications to be overlapped with backpropagation computations. This has been commonly adopted in popular distributed deep learning frameworks. However, there exist two fundamental problems: (1) excessive startup latency proportional to the number of workers for each all-reduce operation; (2) it only achieves sub-optimal training performance due to the dependency and synchronization requirement of the feed-forward computation in the next iteration. We propose a novel scheduling algorithm, DeAR, that decouples the all-reduce primitive into two continuous operations, which overlaps with both backpropagation and feed-forward computations without extra communications. We further design a practical tensor fusion algorithm to improve the training performance. Experimental results with five popular models show that DeAR achieves up to 83\% and 15\% training speedup over the state-of-the-art solutions on a 64-GPU cluster with 10Gb/s Ethernet and 100Gb/s InfiniBand interconnects, respectively. 
\end{abstract}

\section{Introduction}\label{sec:introduction}
Training a complex deep neural network (DNN) model over a large data set requires a massive amount of compute resources and is typically performed on a cluster of GPU machines~\cite{dean2012large,jia2018highly,you2020large}. To accelerate distributed training, many different ways of parallelism have been proposed recently, such as data-parallel~\cite{goyal2017accurate}, model-parallel~\cite{dean2012large}, pipeline-parallel~\cite{huang2019gpipe}, and the combination of the above~\cite{narayanan2021efficient}. Among them, the data-parallel synchronous stochastic gradient descent (S-SGD) algorithms are the most popular when each worker machine has sufficient GPU memory to hold the training model. In S-SGD, the training data is sharded across multiple GPU workers. Each worker iteratively updates the training model by aggregating the local gradients computed with local data samples. To efficiently support gradient aggregation, current training frameworks use the all-reduce architecture~\cite{awan2016efficient,goyal2017accurate,jia2018highly,cho2019blueconnect,chu2020nv,shi2020quantitative}, in which gradient aggregation is performed with an all-reduce collective. The all-reduce architecture has been widely adopted in practice to distributed training, according to the MLPerf training benchmarks\footnote{\url{https://mlcommons.org/en/training-normal-11/}}.

As the model size and the number of workers increase, gradient aggregation requires extensive data communications, which easily become the bottleneck~\cite{xu2017performance,shi2018performance}.
System-level optimizations are thus needed to address this scalability issue. One effective approach that exploits the layer-wise structure of DNN models is to pipeline gradient calculation (computing tasks) with gradient aggregation (communication tasks) in the backpropagation stage, so as to hide the communication overhead and thus improve the system throughput~\cite{zhang2017poseidon,awan2017s}. This approach, known as wait-free backpropagation (WFBP)~\cite{zhang2017poseidon}, has been implemented as the default mechanism in modern deep learning (DL) frameworks such as TensorFlow, PyTorch-DDP~\cite{pytorchddp}, and Horovod~\cite{sergeev2018horovod,romero2022accelerating}. However, WFBP only pipelines communications with gradient computations in the backpropagation stage, which does not consider the feed-forward stage, thus making it sub-optimal. It is worth pointing out that feed-forward computations account for around one third of the total computation time in each iteration~\cite{sze2017efficient}, which can be properly exploited to further accelerating the training speed. 

However, it is challenging to enable pipelining between the communication tasks for gradient aggregation and the next iteration's feed-forward computing tasks under the all-reduce architecture, for two reasons. First, a tensor's gradient aggregation is an all-reduce primitive, which can only begin after its gradient has been calculated in backpropagation and should be synchronized before the next iteration's feed-forward computation. Thus, it only allows coarse-grained scheduling between communications and computations. Second, the all-reduce communication tasks are coming in a first-in, first-out (FIFO) order with the dependency of backpropagation computing tasks. Communication tasks can be re-ordered to be pipelined with feed-forward computing tasks. Yet, different workers execute the computing tasks concurrently, such a re-ordering needs to be done collectively in a consistent manner by all workers to ensure the correctness of all-reduce results. Therefore, this requires synchronization among workers in each iteration, which causes extra communication overheads. 

To address the two challenges above, we propose a new scheduling algorithm called DeAR\footnote{Source code can be found in \url{https://github.com/lzhangbv/dear_pytorch}.} that \underline{de}couples the \underline{a}ll-\underline{r}educe primitive to two operations, so as to enable fine-grained scheduling without introducing extra communication overhead. DeAR applies three novel techniques to distributed training for the all-reduce architecture. To the best of our knowledge, we are the first to decouple the all-reduce primitive without introducing extra time costs so that communications become possible to be overlapped with feed-forward computations in distributed training. 

First, though the all-reduce operation is a primitive in distributed training, all-reduce implementations can be handled as a combination of basic routines~\cite{barnett1994interprocessor,rabenseifner2004optimization,thakur2005optimization,hoefler2010toward}. For example, a classic implementation of the widely used ring-based all-reduce is a combination of a reduce-scatter collective followed by an all-gather collective~\cite{barnett1994interprocessor,rabenseifner2004optimization}. Based on the nature of all-reduce implementations, we decouple the all-reduce primitive to two continuous collectives in distributed training, which allows a fine-grained schedule of communication tasks. 

Second, given that one all-reduce primitive is decoupled into two operations, we propose to schedule the first operation to be pipelined with backpropagation computing tasks, and the second operation pipelined with feed-forward computing tasks. By doing so, there is no need to re-order the communication tasks while enabling the pipelining between the communication tasks and all the computing tasks without introducing any extra communication overhead during training. 

Third, due to the pipelining between the communication tasks and feed-forward computing tasks, tensor fusion techniques~\cite{sergeev2018horovod,shi2019mg,shi2021mg}, which have been proven effective in reducing the latency overhead in WFBP~\cite{zhang2017poseidon}, becomes impractical in DeAR. The main challenge is how to determine which tensor should (not) be fused. To this end, we propose a dynamic tensor fusion algorithm using Bayesian optimization in DeAR to judiciously determine which tensors should be fused to improve the training efficiency, without any prior knowledge about the model and cluster configurations.

We implemented DeAR atop PyTorch. Our implementation provides an easy-to-use API such that users can integrate our training algorithm by adding a few lines of code. Extensive experiments are conducted with popular DNNs on a 64-GPU cluster under various system configurations. Experimental results show that, compared with the state-of-the-art solutions, including PyTorch-DDP, Horovod, MG-WFBP~\cite{shi2019mg}, and ByteScheduler~\cite{peng2019generic}, DeAR accelerates the model training by up to 83\% and 15\% over 10Gb/s Ethernet and 100Gb/s InfiniBand interconnects, respectively. In all experiments, the training speedup enabled by DeAR reaches 72.3-99.2\% of the maximum possible.

\section{Background and Motivation}\label{sec:background}

\subsection{Mini-batch SGD}
The training of DNN models is to minimize a designed loss function $\mathcal{L}(w, X)$, where $w$ is the model parameter and $X$ is the training data. In mini-batch SGD, the model parameters is updated iteratively based on its first-order gradient. Specifically, at each iteration $i$, a mini-batch data ($X_i$) is randomly sampled to calculate the loss through feed-forward from the first layer to the last layer; and then the first-order gradient w.r.t. the model parameter is calculated through backpropagation. Then, the gradient is used to update the parameter. Formally, the update formula at the $i^{th}$ iteration can be represented as follows.
\begin{equation}
    w_{i+1}=w_i-\eta \nabla \mathcal{L}(w_i, X_i),
\end{equation}
where $\eta$ is the learning rate, $w_i$ and $X_i$ are the model parameter and sampled data at iteration $i$, respectively. Thus, in a single-GPU environment, the training time is mainly consumed in the feed-forward and backpropagation computing tasks.

\subsection{S-SGD}
When exploiting multiple workers (e.g., GPUs) to train a single model, synchronous SGD (S-SGD) with data parallelism is a de-facto approach for training as it preserves the convergence properties of mini-batch SGD. In S-SGD, each iteration's training data $X_i$ is distributed to $P$ workers as $X_i^p$ at worker $p$, where $p=1,2,...,P$ on a $P$-worker cluster, and all workers keep consistent parameters at every iteration. The update rule of S-SGD is
\begin{equation}
    w_{i+1}=w_i-\eta \frac{1}{P}\sum_{p=1}^{P}\nabla \mathcal{L}(w_i, X_i^p).
\end{equation}
It is seen that the distributed gradients should be aggregated before updating the model parameter, which introduces communication costs and limits the system scaling efficiency. In practice, the gradient aggregation (GA) can be implemented through a parameter server~\cite{li2014scaling} or an all-reduce collective. We focus on the all-reduce implementation in this work. In summary, the iteration time of S-SGD contains the feed-forward computation time, the backpropagation computation time, and the communication time of gradient aggregation. 

\begin{figure}[!t]
	\centering
	\begin{subfigure}{0.48\textwidth}
		\includegraphics[width=\linewidth]{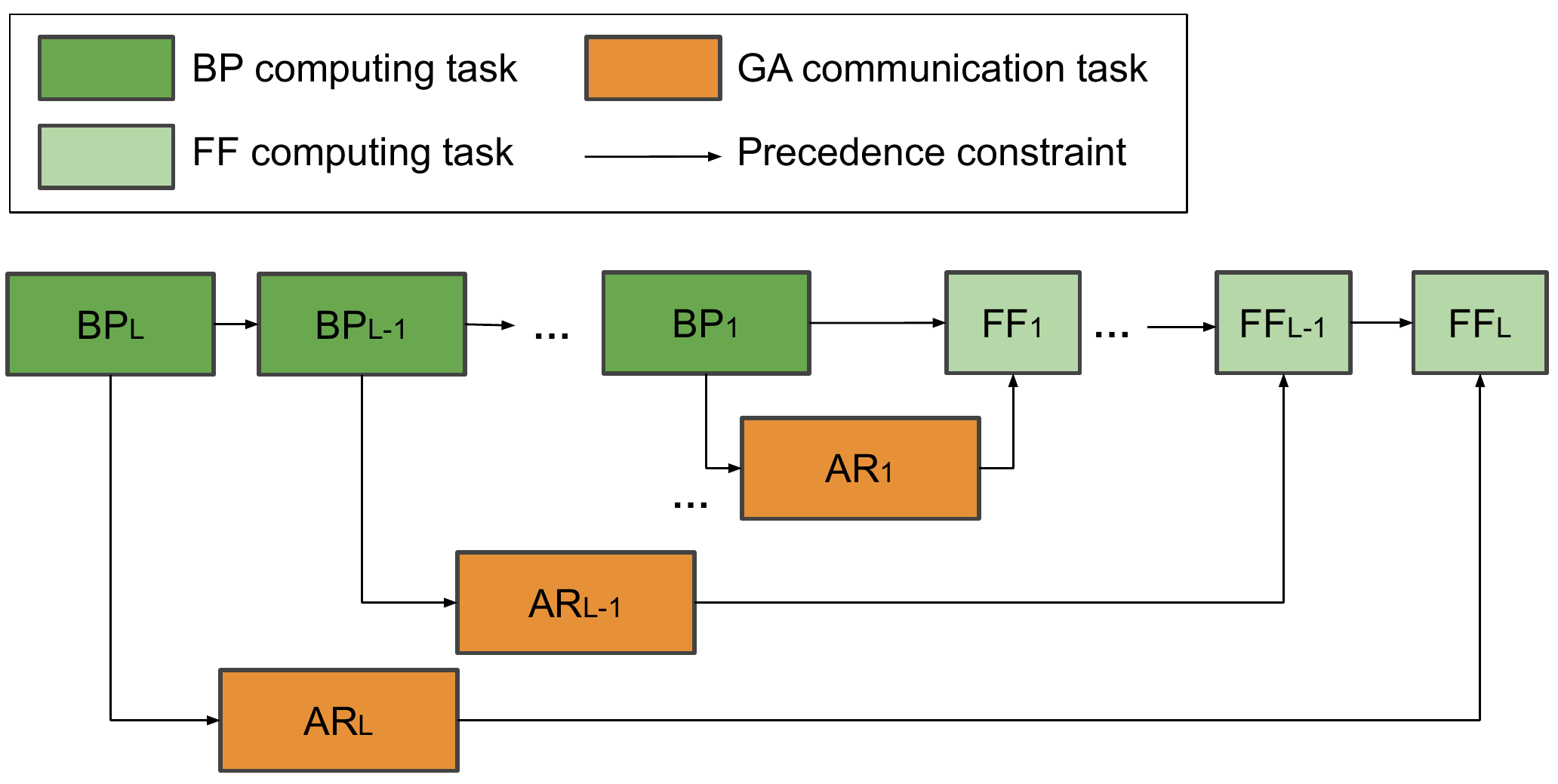}
		\caption{DAG of computing and communication tasks.}
	\end{subfigure}
	\begin{subfigure}{0.48\textwidth}
	\vspace{5pt}
		\includegraphics[width=\linewidth]{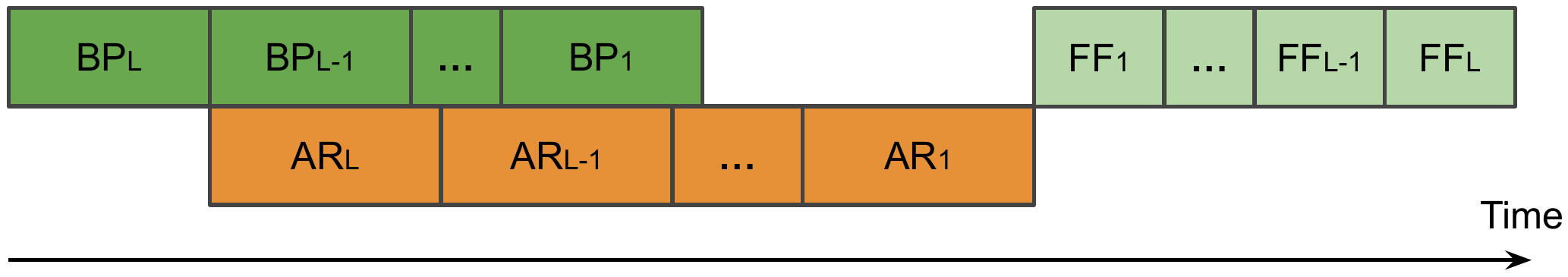}
		\caption{WFBP.}
	\end{subfigure}
	\begin{subfigure}{0.48\textwidth}
	\vspace{5pt}
		\includegraphics[width=\linewidth]{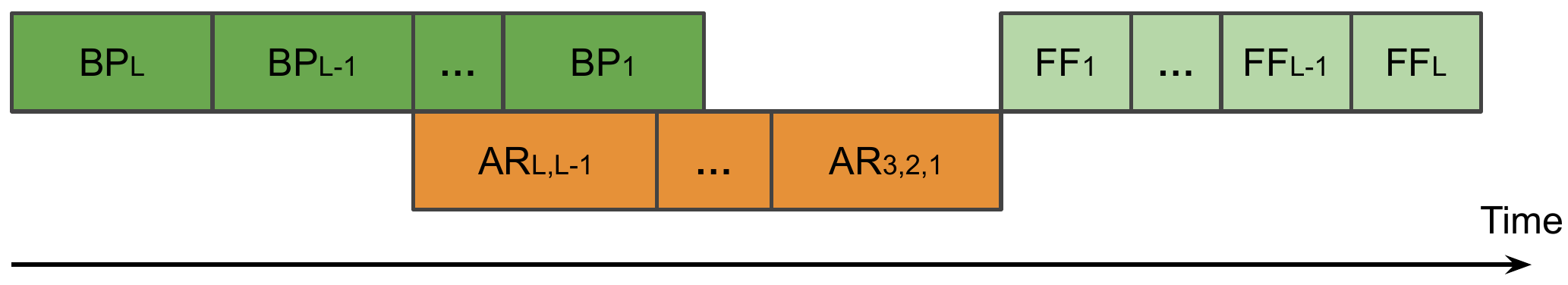}
		\caption{WFBP with tensor fusion.}
	\end{subfigure}
	\begin{subfigure}{0.48\textwidth}
	\vspace{5pt}
		\includegraphics[width=\linewidth]{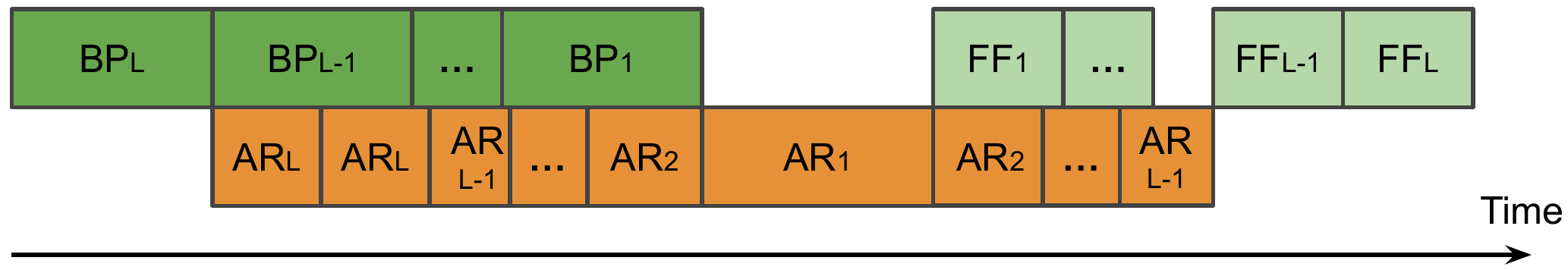}
		\caption{ByteScheduler: priority scheduling and tensor partitioning.}
	\end{subfigure}
    	\caption{(a) The DAG of computing and communication tasks in an $L$-layer DNN, and (b-d) the timeline of S-SGD algorithms with different schedules. (b) WFBP: Gradient communication of each layer begins after that layer's gradients are calculated and the communications are executed in a FIFO order. (c) The gradients of nearby layers are fused to be communicated together. (d) ByteScheduler: large tensors may be partitioned into multiple smaller tensors and the order of communications is based on their priorities but not FIFO.}
	\label{fig:sgd-dag}
\end{figure}

Due to the layer-wise structure of DNN models, the computing tasks and communication tasks can be organized as a directed acyclic graph (DAG) as shown in Fig.~\ref{fig:sgd-dag}(a). One layer's communication ($\text{AR}_{l}$) can only begin after its gradient has been calculated ($\text{BP}_{l}$), and its feed-forward computation ($\text{FF}_{l}$) should wait for the completion of $\text{AR}_{l}$. According to the DAG, it is possible to schedule the order of different tasks so that they can be overlapped to shorten the iteration time. 

\subsection{Wait-free backpropagation}
Gradient aggregation of some layers can be overlapped with backpropagation using the wait-free backpropagation algorithm (WFBP)~\cite{zhang2017poseidon,awan2017s}, in which the gradient communication can immediately begin after the gradient is calculated. Due to the nature of backpropagation, where the gradients are calculated from the last layer to the first layer, multiple layers communications are scheduled with a first-in-first-out (FIFO) order as shown in Fig.~\ref{fig:sgd-dag}(b). In modern DNN models, there are many layers which have a relatively small number of gradients that need to be aggregated, thus WFBP requires tensor fusion (e.g., MG-WFBP~\cite{shi2019mg}) to alleviate the startup overhead in all-reduce communications as shown in Fig.~\ref{fig:sgd-dag}(c). 

However, WFBP and its variant only allow the gradient aggregation communication tasks to be pipelined with backpropagation computing tasks as shown in Fig.~\ref{fig:sgd-dag}(b)(c). That is, the feed-forward computing tasks of the next iteration can only begin after all the GA communication tasks of the current iteration have been completed. Thus, the communication tasks have no opportunity to be pipelined with the feed-forward computing tasks, which is sub-optimal if the communication time cannot be fully hidden by the backpropagation computation time. The feed-forward computing tasks normally consume around one to third of the total computation time at each iteration~\cite{sze2017efficient}. If one can pipeline the communication tasks with feed-forward computing tasks, the one to third computation time could also be saved.

\subsection{Priority scheduling and tensor partitioning}
Though ByteScheduler regards the gradient aggregation as a pair of (PUSH, PULL) in the PS architecture to enable a finer-gained schedule, it cannot use the (PUSH, PULL) feature in all-reduce which is a primitive in existing deep learning frameworks. Instead, to enable some communication tasks to be overlapped with feed-forward computing tasks, ByteScheduler~\cite{peng2019generic} re-orders the communication tasks and issues the tasks in an ``optimal'' order by allowing large tensors to be partitioned into multiple small tensors as shown in Fig.~\ref{fig:sgd-dag}(d). First, the communication of the second layer (i.e., AR$_2$), which can only begin after AR$_L$ to AR$_3$ in a FIFO order, is scheduled to be executed prior to AR$_{L-1}$. Second, some large tensors may be partitioned into multiple small tensors to provide finer-grained scheduling. For example, the tensor of layer 2's gradient is partitioned to two tensors which can be separately completed with two all-reduce operations. The priority scheduling and tensor partitioning techniques in enabling the communication tasks to be pipelined with feed-forward computing tasks may work well in the PS architecture~\cite{peng2019generic}, but it would have significant performance issues in the all-to-all architecture due to the following two problems.

First, re-ordering the communication tasks requires all workers, which execute computing tasks concurrently during training, to have a consensus on communicating a particular tensor. In other words, before aggregating the gradient of a layer, all workers should negotiate with each other that the layer is ready for communicating to ensure the correctness of training. This would introduce extra communication overheads. Even though the negotiation only needs to communicate several bytes of data, it may have significant latency with the increasing number of workers, especially on high-latency interconnects (e.g., 10Gb/s Ethernet).

Second, using tensor partitioning for a finer-grained schedule may introduce extra startup overheads of communications. Generally, the time cost of an all-reduce communication contains a startup overhead that is proportional to the number of workers~\cite{rabenseifner2004optimization,thakur2005optimization,hoefler2010toward,shi2019mg}. For example, in the widely used ring-based all-reduce algorithm, which is a default in NCCL, the startup time is linear to the number of GPUs~\cite{thakur2005optimization}. Therefore, partitioning a tensor to $n$ smaller tensors to be communicated separately would introduce extra $n-1$ startup overheads. For example, on a 64-GPU cluster with 10Gb/s Ethernet, all-reducing a 1MB message takes around 4.5ms, while all-reducing a 500KB message takes around 3.9ms.

In summary, existing scheduling techniques to enable the pipelining between communication tasks and feed-forward computing tasks are impractical for distributed training in the all-to-all architecture. Pipelining the communication tasks with feed-forward computing tasks is expected to save one to third of the computation time, but the introduced extra communication overhead in ByteScheduler may be larger than the hidden computation time, resulting an even worse performance.

This motivates us to decouple the all-reduce primitive based on its implementation nature to two operations for a finer-grained schedule, and it does not introduce any extra communication overhead.

\section{DeAR: Decoupling the all-reduce primitive}\label{sec:design}
The design philosophy of our DeAR is to decouple the all-reduce primitive to two continuous operations without introducing extra communication overheads. 

\subsection{Decoupling all-reduce with zero overhead}
According to the inherent feature of all-reduce primitive that tries to maximally utilize the network bandwidth or minimally reduce the latency~\cite{thakur2005optimization,patarasuk2009bandwidth}, it should be implemented with multiple rounds of communications, each of which has multiple workers participating in sending and receiving messages simultaneously. Thus, it is very common that the all-reduce algorithm is implemented with a combination of other basic routines~\cite{rabenseifner2004optimization}. For example, the ring-based all-reduce algorithm can be implemented by a ring-based reduce-scatter operation followed by a ring-based all-gather operation~\cite{thakur2005optimization}. Thus, theoretically, the all-reduce primitive can be decoupled into two or more continuous operations whose total time equals to the time cost of the all-reduce primitive. The decoupled operations of a primitive will allow finer-grained tasks scheduling in distributed training.

\begin{figure}[!t]
	\center
	\begin{subfigure}{0.48\textwidth}
		\includegraphics[width=\linewidth]{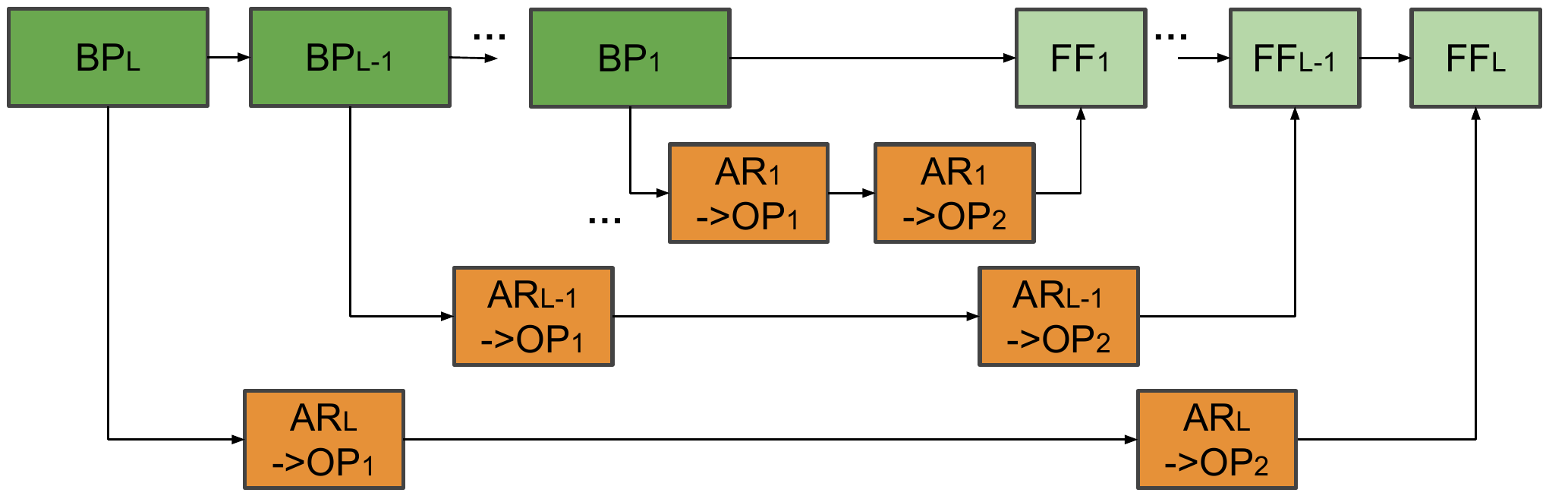}
		\caption{DAG of computing and communication tasks in DeAR.}
	\end{subfigure}
	\begin{subfigure}{0.48\textwidth}
	\vspace{5pt}
		\includegraphics[width=\linewidth]{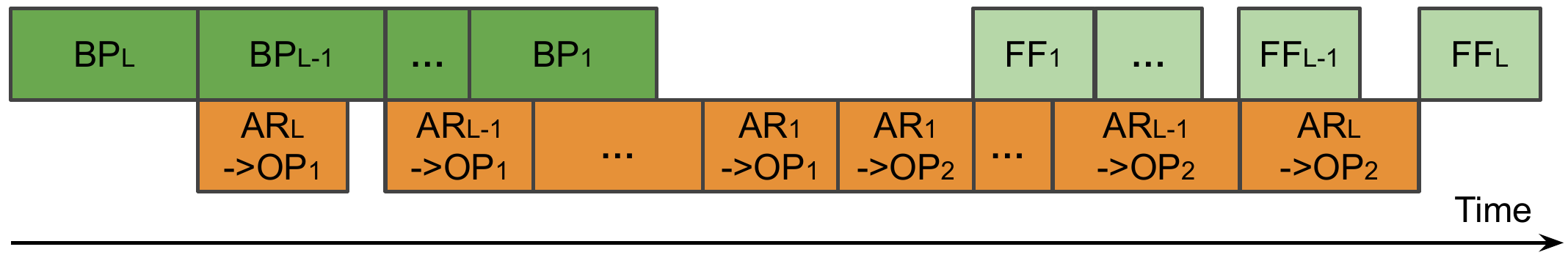}
		\caption{Timeline of DeAR with an order of FIFO for communications.}
	\end{subfigure}
	\begin{subfigure}{0.48\textwidth}
	\vspace{5pt}
		\includegraphics[width=\linewidth]{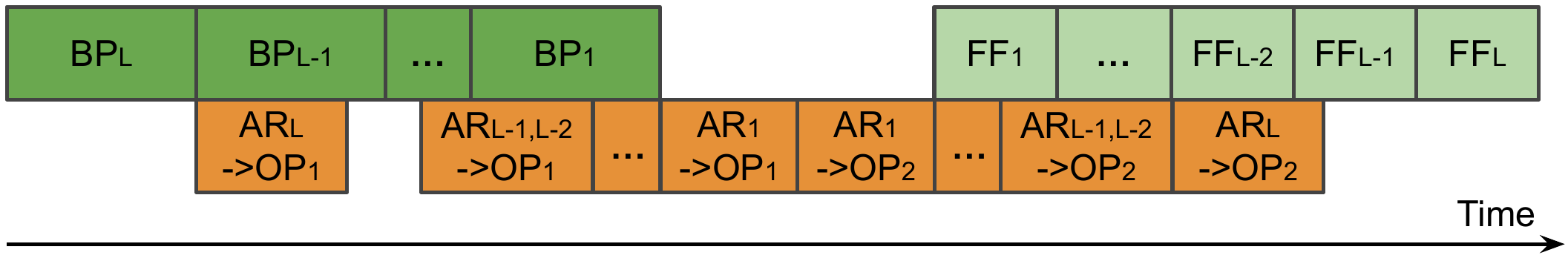}
		\caption{Timeline of DeAR with tensor fusion.}
	\end{subfigure}
	\caption{(a) The DAG of computing and communication tasks with an $L$-layer DNN in DeAR. (b) DeAR without tensor fusion: the decomposed communications are executed in a FIFO order. (c) DeAR with tensor fusion: Nearby gradients could be merged to a single one for the decoupled operations.}
	\label{fig:DeAR}
\end{figure}

As our goal is to enable some communication tasks to be pipelined with feed-forward computing tasks, we break down the all-reduce operation $OP_{ar}$ into two continuous communication operations, say $OP_1$ and $OP_2$. Note that the total time of $OP_1$ and $OP_2$ equals to the time of $OP_{ar}$, which means the decoupling is free. Therefore, $OP_1$ of different layers can still be pipelined with backpropagation computing tasks, while $OP_2$ can be pipelined with feed-forward computing tasks. The DAG of computing and communication tasks in DeAR is shown in Fig.~\ref{fig:DeAR}(a). One layer's gradient aggregation is composed of two continuous communication operations. Compared to Fig.~\ref{fig:sgd-dag}(a), the fine-grained DAG with decoupled all-reduce allows us to schedule $OP_1$ and $OP_2$ communication tasks separately, which offers great opportunities to pipeline the communication tasks with feed-forward computing tasks without tensor partitioning. 

In this work, we use the ring-based all-reduce algorithm, which is widely used in distributed training, as an example to show how we decouple it with zero overhead. Note that the key idea of DeAR can be applied in any all-reduce algorithms as long as they can be decoupled into two operations without introducing any extra overhead. In the ring-based algorithm on a $P$-worker cluster, the $d$ elements are divided to $P$ chunks, each of which has $d/P$ elements. In the first step, each chunk will be reduced to each worker via $P-1$ communication rounds, which is a reduce-scatter operation and it takes a time complexity of
\begin{equation}\label{equ:time-reducescatter}
   t_{rs} = (P-1)(\alpha+\frac{d}{P}\beta),
\end{equation}
where $\alpha$ and $\beta$ are the latency and transmission time per element between two workers according to the $\alpha-\beta$ cost model~\cite{thakur2003improving}. As we only focus on the communication time, we omit the overhead of arithmetic operations of accumulating elements in Eq.~\ref{equ:time-reducescatter}.

In the second step, each reduced chunk at every worker is broadcast to all other workers, which is an all-gather operation and it also takes $P-1$ communication rounds in the ring-based algorithm. The all-gather operation has a time complexity of
\begin{equation}\label{equ:time-allgather}
    t_{ag} = (P-1)(\alpha+\frac{d}{P}\beta).
\end{equation}

Putting Eq.~\ref{equ:time-reducescatter} and Eq.~\ref{equ:time-allgather} together, we achieve the time complexity of an all-reduce operation as follows.
\begin{equation}\label{equ:time-allreduce}
    t_{ar} = 2(P-1)\alpha+\frac{2(P-1)d}{P}\beta.
\end{equation}

\subsection{Pipelining communication tasks without re-ordering}
By decoupling the all-reduce primitive to two continuous operations, it becomes possible to pipeline the first communication operation with backpropagation computing tasks, and pipeline the second communication operation with feed-forward computing tasks as shown in~\ref{fig:DeAR}(b). To guarantee data dependencies between tasks at run-time, we propose 1) BackPipe: starting the communication task of $OP_1$ immediately when the gradient of one layer is ready in the backward pass, and 2) FeedPipe: waiting for the completion of the communication task of $OP_2$ of one layer before its feed-forward computation, and starting the communication task of $OP_2$ of the next layer. Besides, we synchronize all the tasks of $OP_1$ at the end of BackPipe to ensure the dependencies between $OP_1$ and $OP_2$.  

In doing so, our DeAR can execute the communication tasks asynchronously to support pipelining with both feed-forward and backpropagation computing tasks, while preserving data dependencies between tasks without any requirement to adjust the order of communication tasks. That is, communication tasks are issued among all workers consistently from the last layer to the first layer during backpropagation and its reverse order during feed-forward, respectively. Therefore, all workers do not need the time-consuming negotiation with each other to reach a consensus in which tensors should be aggregated.

In summary, compared with the WFBP~\cite{zhang2017poseidon,awan2017s} or its variant~\cite{shi2019mg,sergeev2018horovod} algorithms, DeAR is able to overlap the gradient aggregation communications with both feed-forward and backpropagation computing tasks. Compared with ByteScheduler~\cite{peng2019generic}, DeAR enables a finer-grained tasks schedule in distributed training without the requirement of partitioning tensors and re-ordering the communication tasks. 

Moreover, DeAR reserves the property of tensor fusion as like WFBP, where the gradients in nearby layers can be merged together to be communicated once to reduce the startup overhead. Unlike ByteScheduler which exploits tensor partitioning (a mutual operation with tensor fusion) to provide a fine-grained schedule, DeAR schedules $OP_1$ with backpropagation computing tasks and $OP_2$ with feed-forward computing tasks, which means $OP_1$ in different layers are possible to be merged to be communicated together and it is similar to $OP_2$. We discuss the details about tensor fusion in the following section.

\section{Tensor fusion in DeAR}\label{sec:tensorfusion}
Tensor fusion~\cite{sergeev2018horovod,shi2019mg} has been proven to be a simple yet effective approach to reducing the startup overheads of all-reduce operations. It has become a default feature in distributed DL frameworks like PyTorch-DDP~\cite{pytorchddp} and Horovod~\cite{sergeev2018horovod}. However, how to determine which layers should be merged is quite challenging as merging any nearby layers requires to wait for their backpropagation computing tasks to be completed.

\subsection{Preliminary of tensor fusion techniques}
In the standard all-reduce primitive case (like PyTorch-DDP and Horovod), where the communication tasks only overlap with backpropagation computing tasks, a buffer with a pre-defined size (e.g., 25MB in PyTorch-DDP and 64MB in Horovod) is allocated to store the ready-to-communicate tensors. When the total size of ready tensors reaches out of the buffer size, the buffer is communicated for aggregation with an all-reduce operation, which means multiple tensors stored in the buffer only communicate once at each iteration. The buffer data is then copied back to the original tensors when the current all-reduce operation completes. Due to the gradients in different layers become ready in a backward order, fusing any two layers needs to wait for the completion of all gradients in these two layers as shown in Fig.~\ref{fig:sgd-dag}(c). Thus, it is non-trivial to determine the optimal buffer size to achieve minimal iteration time.

One can also measure the backpropagation computation time of each layer and estimate the communication time of all-reduce to dynamically determine whether the benefit of merging any two nearby layers is larger than the sacrifice of the waiting time~\cite{shi2019mg}. Yet, two main issues may make the solution impractical~\cite{shi2019mg}. First, the layer-wise backpropagation time is quite difficult to be correctly measured as each layer gradient may be computed asynchronously. Second, the variant tensor sizes in a DNN model make it difficult to predict the communication time accurately by a simple model.

Different from the previous works, DeAR pipelines some communication tasks with feed-forward computing tasks, which means that tensor fusion of any two layers may affect the granularity of feed-forward pipelining as shown in Fig.~\ref{fig:DeAR}(c). Layer $L-1$ and layer $L-2$ are fused to be communicated once using $OP_1$ so that $OP_2$ of these two layers should also be invoked only once and it should be synchronized before the feed-forward computation of layer $L-2$. In this case, $OP_2$ of layer $L-1$ cannot be overlapped with the feed-forward computation of layer $L-2$. As a result, though DeAR reserves the property of tensor fusion, it is non-trivial to determine which layers should be fused to achieve minimal iteration time.

\subsection{Bayesian optimization based tensor fusion}
In DeAR, fusing the gradients of any two nearby layers has two drawbacks. 1) It requires to wait for the completion of the two layers' gradient computations to start the communication of reduce-scatter, which means the current ready layer cannot start communication immediately. 2) It reduces the opportunity of overlapping all-gather of one layer with the feed-forward computation of its previous layer. Thus, one should carefully choose the tensor fusion strategy such that the overall iteration time can be shortened. Due to the difficulty in formulating the tensor fusion problem with heuristic or optimal solutions, we choose to use Bayesian optimization (BO)~\cite{snoek2012practical,package2014bo}, which attempts to find good parameters of an unknown objective function in as few number of trials as possible \cite{peng2019generic,sergeev2018horovod}.

\begin{figure}[!t]
    \centering
    \includegraphics[width=0.7\columnwidth]{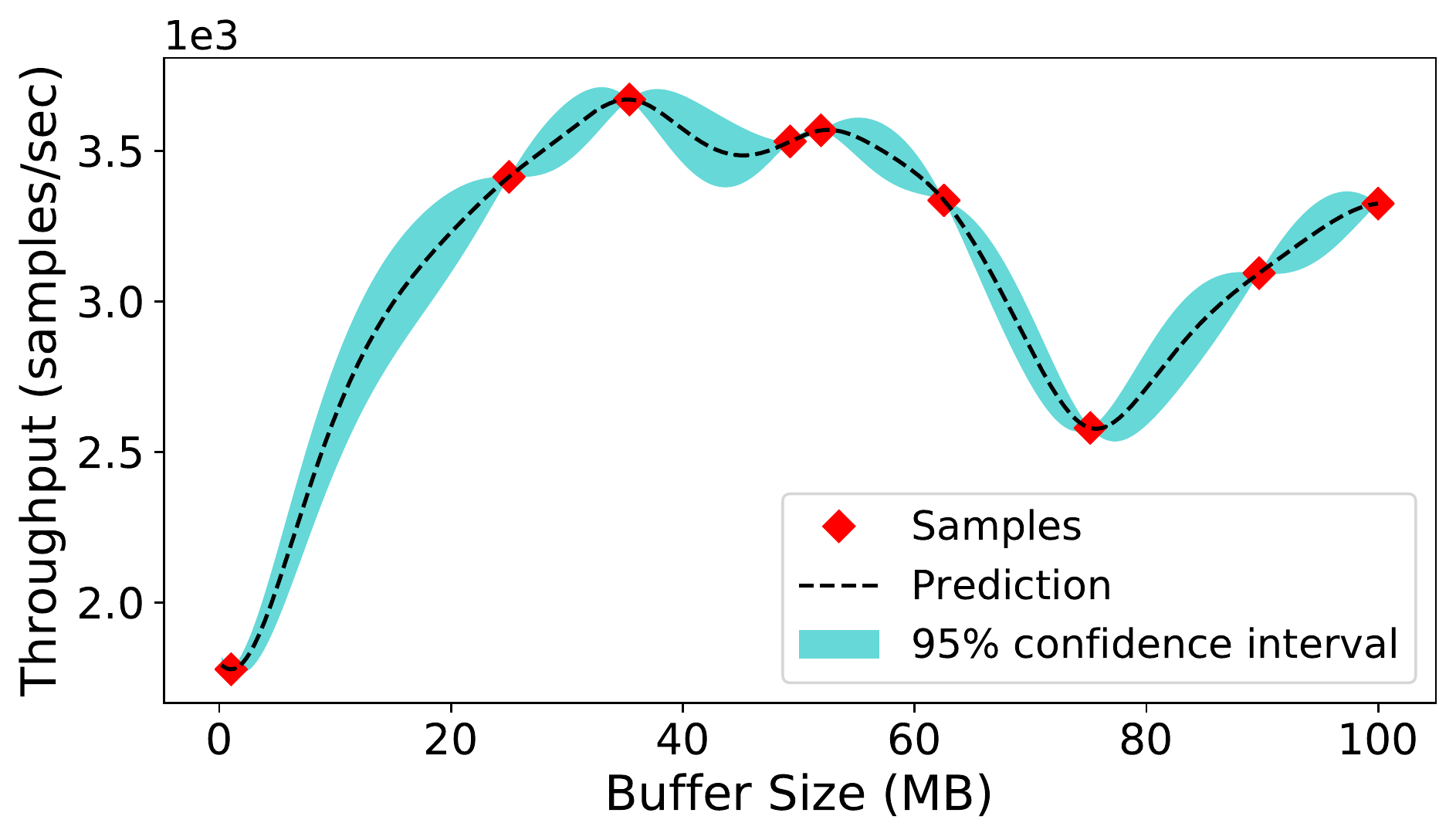}
    \caption{Bayesian optimization example: 9 samples; tuning buffer size for training DenseNet-201. }
    \label{fig:bo-process}
\end{figure}
The target of BO used in DeAR is try to achieve maximum training performance (measured as the system throughput, i.e., the number of training samples that can be processed per second) during run-time in our system. We use $P(x)$ to denote the performance model of our system, which is unknown, and $x$ is the buffer size which is an input parameter used for tensor fusion. Note that different $x$ may generate different tensor fusion solutions. Specifically, nearby layers are put into one group if their total number of gradients does not exceed the size $x$. Gradients in one group will have only one reduce-scatter operation during backpropagation, and one all-gather operation during feed-forward. We would like to update $x$ dynamically such that $P(x)$ converges to a stable value.

BO is effective to find near-optimal tensor fusion solutions for three reasons. First, BO uses the Gaussian process regression in predicting the function value, so it has no constraints on the objective function format and only relies on the existing observations, i.e., $\hat{P}(x_1), \hat{P}(x_2), ..., \hat{P}(x_n)$. Second, BO usually needs a few number of trials to find good solutions, which only requires very small search costs. This is because BO suggests the next system configuration based on a well-defined acquisition function \cite{Alipourfard2017CherryPick}. In this work, we use expected improvement (EI) acquisition function to pick the next point that can maximize the expected improvement over the current best result. Third, BO can tolerate uncertainty with quantitative confidence interval. For example, by tuning the EI hyper-parameter, we find BO can balance between exploitation and exploration during the search process, which is helpful to escape from a local optimum. In general, smaller EI hyper-parameter prefers exploitation (i.e., most points are around the peaks), while larger value prefers exploration (i.e., the points are more spread out across the whole range)~\cite{Alipourfard2017CherryPick}. In this problem, we set EI hyper-parameter as 0.1 to prefer buffer size exploration, e.g., from 1MB to 100MB (see Fig.~\ref{fig:bo-process}).

To support BO during training, we first use a default buffer size $x_1=25$MB to initialize the tensor fusion configuration and measure the average system throughput (i.e., $\hat{P}(x_1)$) over multiple steps (e.g., 10 steps). Based on this measurement $\hat{P}(x_1)$, BO fits the performance function and suggests the next buffer size $x_2$, which can be used to generate a new tensor fusion solution. By repeating this process, BO can predict the performance accurately with enough samples, and find a good tensor fusion solution. For example, in Fig.~\ref{fig:bo-process}, we use BO to find the buffer size for training DenseNet-201~\cite{huang2017densely} in DeAR. With only 9 samples, it returns a nearly optimal value at 35MB with a good confidence. In practice, tens of trials are enough to find a good solution for DeAR (see Figure~\ref{fig:tuning-cost}), while the BO tuner developed for Horovod is much more costly, as it needs to search multiple system configurations including buffer size, cycle time, response cache, and hierarchical collective algorithms~\cite{sergeev2018horovod}.

\section{Implementation}\label{sec:implementation}
\begin{figure}[!t]
	\centering
		\includegraphics[width=0.7\linewidth]{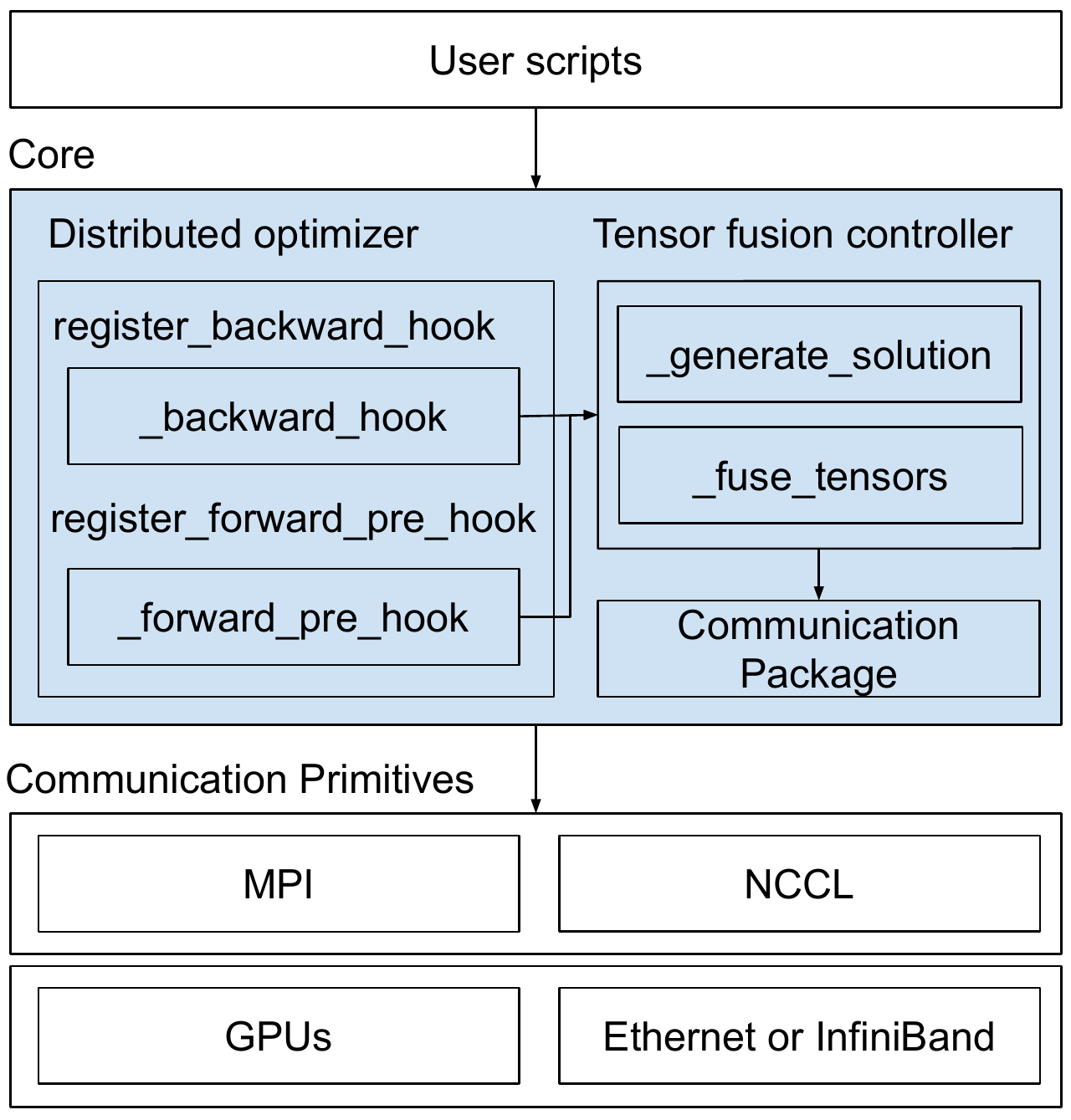}
	\caption{Overview of our system (blue parts are new).}
	\label{fig:design}
\end{figure}
We implement our prototype system, DeAR, based on PyTorch and NCCL. In the system, we wrap a communication library using C/C++ based on NCCL and expose APIs for high-level scripts in Python. The overview of our DeAR implementation is shown in Fig.~\ref{fig:design}. The blue components are new in DeAR, but they are totally transparent to end users. Users only need to change their code (normally with several new inserted lines) to use DeAR. 

\subsection{Workflow of DeAR}
We implement our DeAR as a middle layer between user code and communication primitives. DeAR does not change the original DAG constructed in PyTorch. A \textit{distributed optimizer} is implemented in DeAR to handle the gradient communications in hook functions provided by PyTorch APIs. Before communicating gradients through reduce-scatter or all-gather, gradients should be put in the \textit{tensor fusion controller} which determines whether the pushed tensor should be copied to the buffer to be communicated together. When the buffer should be aggregated among all workers, DeAR invokes the wrapped NCCL APIs (e.g., ncclReduceScatter and ncclAllGather) in \textit{communication package}.

\subsection{User usage}
\begin{lstlisting}[label=samplecode,caption=Code example of using DeAR,language=Python]
import dear # +++
dear.init() # +++
optim = torch.optim.SGD(model.parameters(), ...)
optim = dear.DistOptim(optim, model, ...) # +++
# Training
model.train()
for i in range(epochs):
    for data, target in train_loader:
        train_step(optim, data, target)
        ...
# Validation 
optim.synchronize() # +++
optim.step() # +++
model.eval()
validation()
        
        
\end{lstlisting}

To make our DeAR be easily integrated with existing user code, we design a distributed optimizer (DistOptim) that is exposed to users. Users only need to wrap their original PyTorch optimizer instances to our DistOptim and initialize a new instance of the optimizer as the sample code shown in Listing~\ref{samplecode}. The first two lines should be inserted to initialize the run-time of DeAR. Then line 4 is inserted after the standard optimizer instance and the training code remains unchanged. As DeAR pipelines the communication tasks of current iteration with the next iteration's feed-forward computing tasks, the communication tasks should be forced to synchronize to update the model parameters (lines 12 and 13) before evaluating the model.

\section{Evaluation}\label{sec:evaluation}

\subsection{Experimental settings}
\textbf{Testbeds. } We conduct experiments on a 16-node dense-GPU cluster, which has 64 Nvidia GTX 2080Ti GPUs with four GPUs per node. The cluster is connected with both 10Gb/s Ethernet (10GbE) and 100Gb/s InfiniBand (100GbIB). Thus, we can choose two different network configurations to test the scalability of different algorithms. Each node has 512GiB RAM and the same software configurations. Specifically, each node is installed with Ubuntu18.04, CUDA-10.2, cuDNN-7.6, NCCL-2.10, OpenMPI-4.0, and PyTorch-1.8. We use NCCL APIs for all collective communications in our experiments.

\textbf{DNN models. }We choose two popular types of DNNs. They are image classification models, CNNs, on the ImageNet dataset~\cite{deng2009imagenet}, and NLP pre-training models BERT~\cite{devlin2019bert}. The detailed settings are shown in Table~\ref{table:dnnconfig}. A training sample is an image with a resolution of $224\times 224\times 3$ for CNNs, and a sentence with a length of 64 words for BERTs.

\begin{table}[!ht]
	\centering
		\caption{DNN details for experiments. ``BS'' denotes the mini-batch size per GPU. ``\# Layers'' represents the number of learnable layers. ``\# Tensors'' and ``\# Param.'' denote the number of learnable parameter tensors and the number of elements (million) in these tensors, respectively.}
		\label{table:dnnconfig}
		 \addtolength{\tabcolsep}{-3.4pt}
    \begin{tabular}{|c|c|c|c|c|c|}
    	\hline
        Application & Model & ~BS~ & \# Layers & \# Tensors & \# Param. (M)  \\\hline\hline
        \multirow{3}{*}{\shortstack{Image \\Classification}} & ResNet-50~\cite{he2016deep}& 64 & 107 & 161 & 25.6 \\\cline{2-6}
        
    	& DenseNet-201~\cite{huang2017densely}& 32 & 402 & 604  & 20.0 \\\cline{2-6}
    	
    	& Inception-v4~\cite{szegedy2017inception}& 64 & 299 & 449  & 42.7 \\\cline{1-6}
    	
        \multirow{2}{*}{\shortstack{NLP \\Pre-training}} & BERT-Base~\cite{devlin2019bert}& 64 & 105 & 206  & 110.1  \\\cline{2-6}
        
        & BERT-Large~\cite{devlin2019bert}& 32  & 201 & 398  & 336.2 \\\cline{1-6}
    \end{tabular}
\end{table} 

\textbf{Baselines. }We compare our system with existing state-of-the-art systems including Horovod-0.21.3~\cite{sergeev2018horovod}, PyTorch-DDP~\cite{pytorchddp} (at PyTorch-1.8), ByteScheduler\footnote{\url{https://github.com/bytedance/byteps/tree/bytescheduler/bytescheduler} (at GitHub commit 33fe89). Note that ByteScheduler cannot support PyTorch-1.8, so we configure PyTorch-1.4 when running ByteScheduler.}, and MG-WFBP\footnote{\url{https://github.com/HKBU-HPML/MG-WFBP} (at GitHub commit 5b8ad5)}. All the systems are based on the DL framework PyTorch.

\subsection{Verification of all-reduce breakdowns}
\begin{figure}[!t]
	\centering
	\hspace{-10pt}
	\subfloat[Small size: (1K, 1M)]{
        \includegraphics[width=0.47\columnwidth]{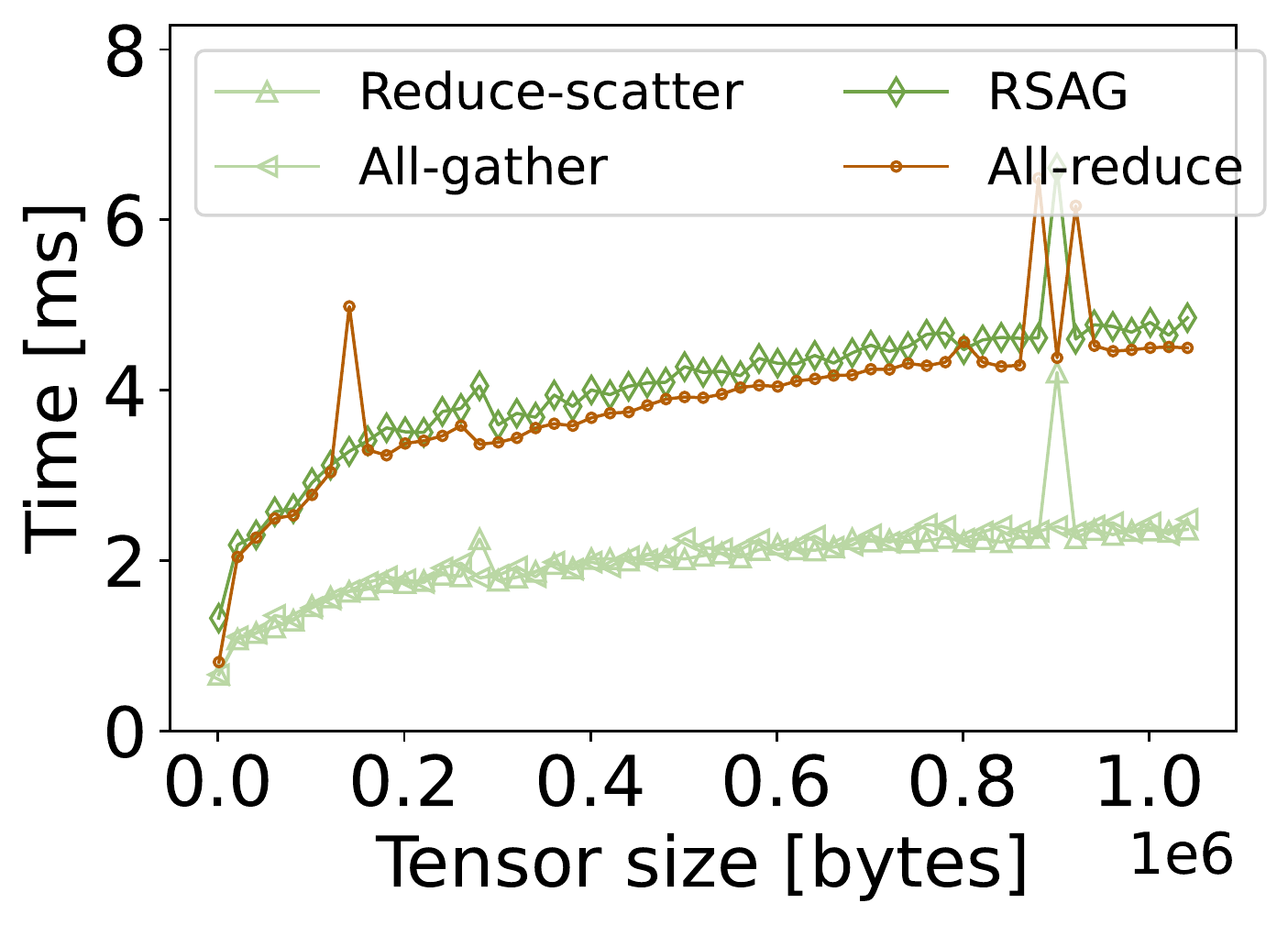}
    }
    \subfloat[Large size: (1M, 100M)]{
        \includegraphics[width=0.49\columnwidth]{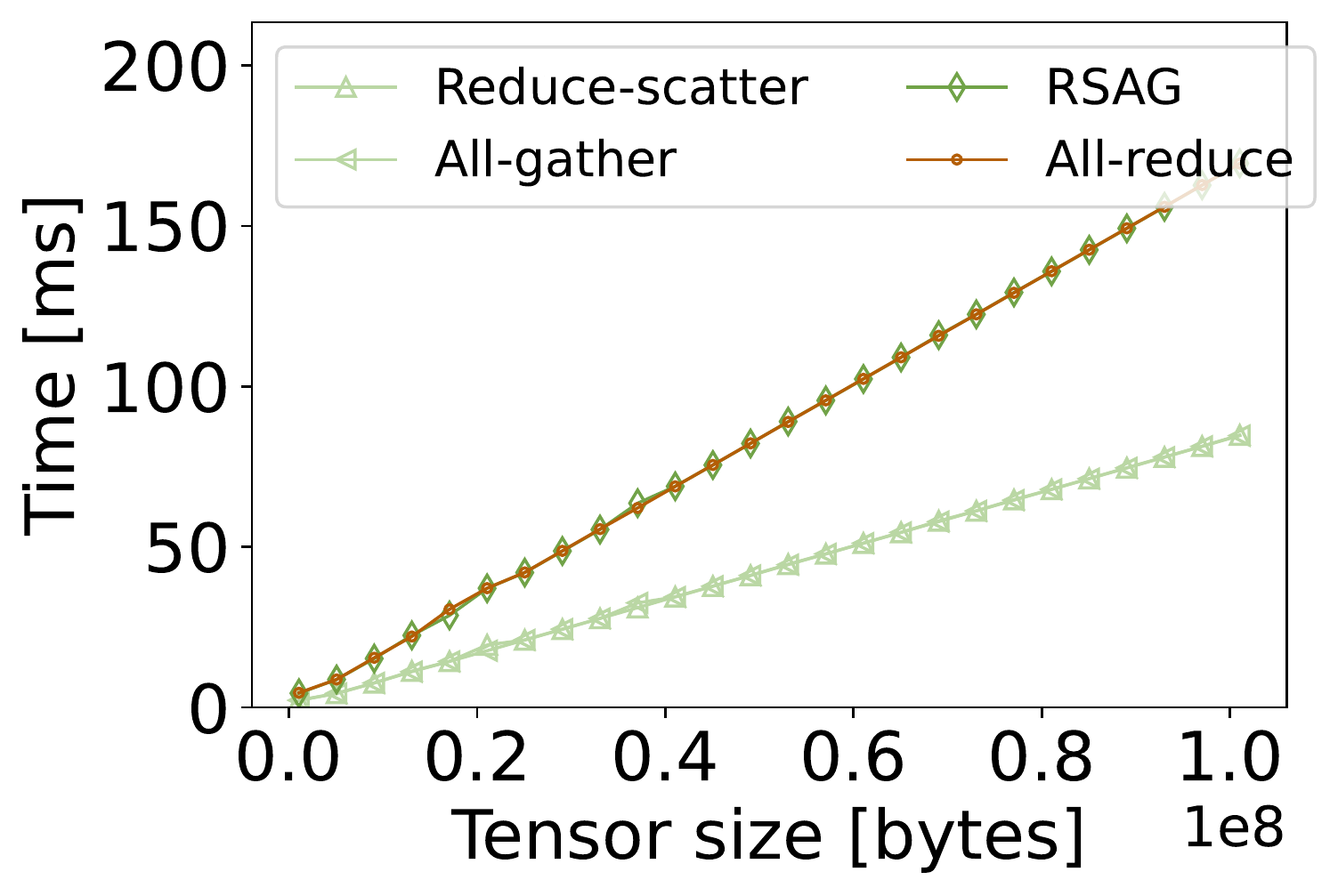}
        \vspace{2pt}
    }
	\caption{Performance comparison with different message aggregation methods. ``RSAG'' represents the all-reduce algorithm that is implemented with a reduce-scatter operation followed by an all-gather operation.}
	\label{fig:allreducecompr}
\end{figure}
To verify that the decoupling of all-reduce has almost zero overhead with different message sizes on dense-GPU clusters, we measure the elapsed-time of all-reduce and its decoupling methods (i.e., reduce-scatter, all-gather, and their combination). We run experiments using nccl-tests\footnote{\url{https://github.com/NVIDIA/nccl-tests}} on the 64-GPU cluster connected with 10GbE. The results are shown in Fig.~\ref{fig:allreducecompr}, in which we can see that both reduce-scatter and all-gather take around half of the time of all-reduce with both small and large sizes messages. Thus, DeAR enables a finer-grained schedule for the decoupled communication tasks without introducing extra communication overheads.

\begin{figure}[!t]
    \captionsetup[subfigure]{justification=centering}
	\centering
	\subfloat[64 GPUs with 10GbE]{
        \includegraphics[width=0.9\columnwidth]{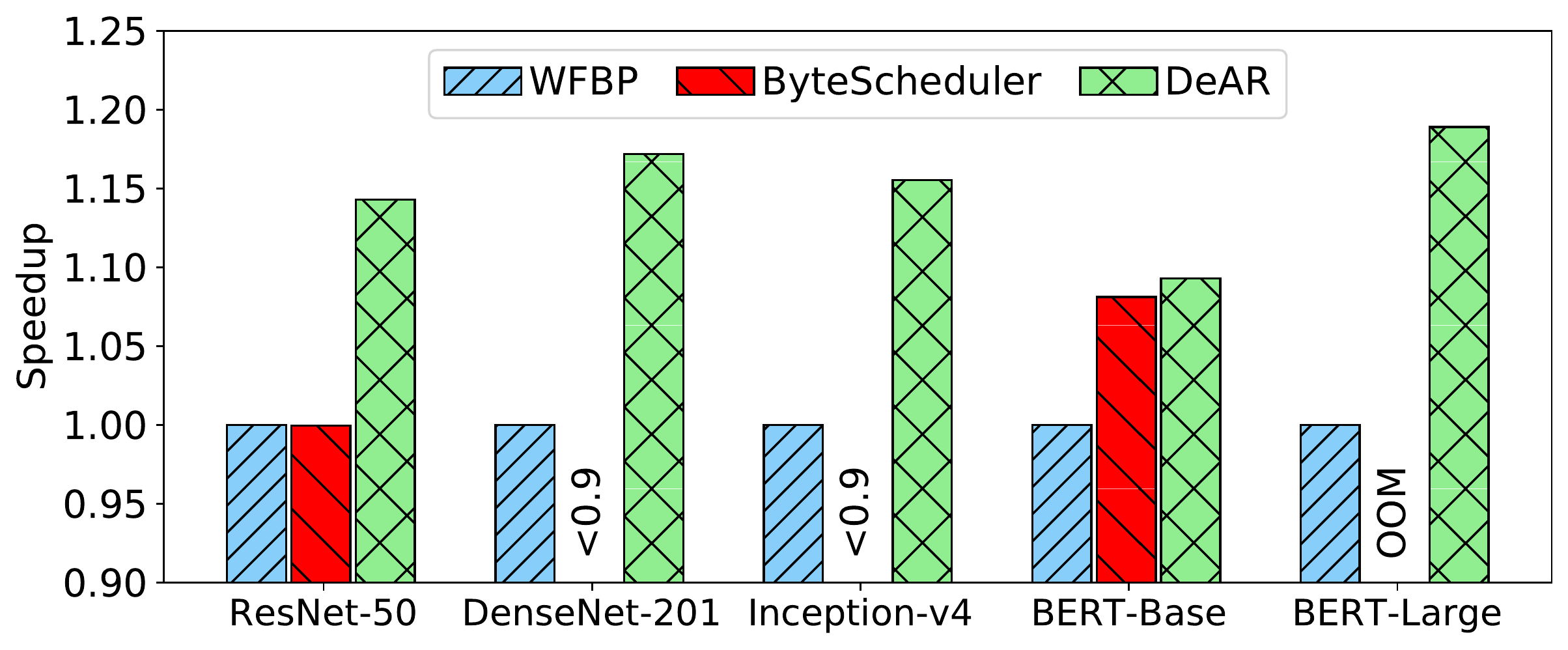}
    }\\
	\subfloat[64 GPUs with 100GbIB]{
        \includegraphics[width=0.9\columnwidth]{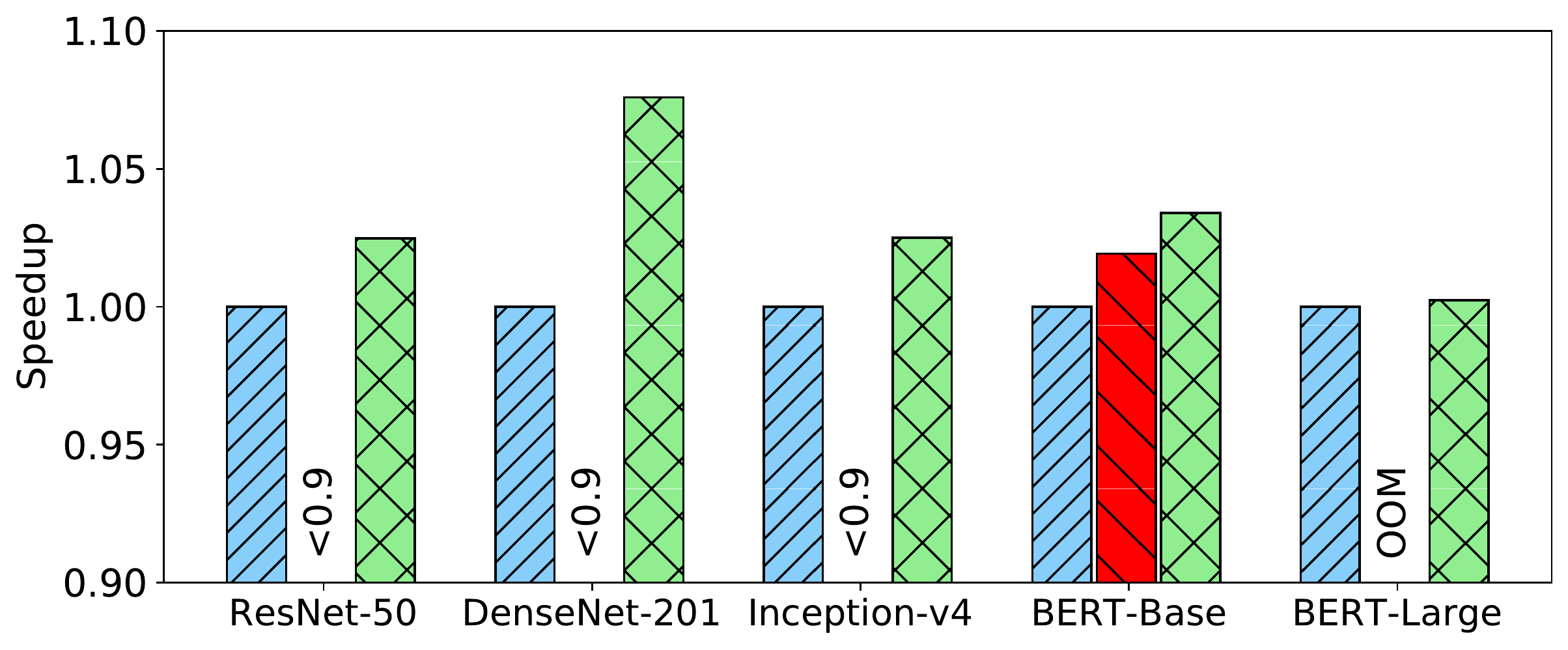}
    }
	\caption{Speedups without tensor fusion. The performance of WFBP is used as the baseline. ByteScheduler runs out-of-memory (OOM) in BERT-Large.}
	\label{fig:speed-tf0}
\end{figure}
\subsection{Speed comparison w/o tensor fusion}

We first show the benefits of overlapping the communication tasks with feed-forward computing tasks in our DeAR without any tensor fusion techniques. We compare DeAR with existing scheduling algorithms without tensor fusion including WFBP~\cite{zhang2017poseidon} and ByteScheduler~\cite{peng2019generic}. For a fair comparison with WFBP, we implement the all-reduce API with a reduce-scatter operation followed by an all-gather operation. The speedup results are shown in Fig.~\ref{fig:speed-tf0}, using the performance of WFBP as the baseline.

Compared with WFBP which only pipelines the communications with backpropagation computing tasks, our DeAR achieves 6\%-19\% improvement in all tested cases due to our fine-grained schedule where the communication tasks are pipelined with both feed-forward and backpropagation computing tasks.

ByteScheduler also pipelines the communications with both feed-forward and backpropagation computing tasks, but it uses tensor partitioning and task re-ordering to achieve finer-grained scheduling of tasks. However, tensor partitioning and re-ordering require extensive extra communication overheads under the all-to-all architecture, ByteScheduler runs very slow in most cases especially on CNNs, thus its bars in Fig.~\ref{fig:speed-tf0} are very low (e.g., $< 0.9$). DeAR significantly outperforms ByteScheduler, especially on CNNs. In contrast, on BERT models which have much larger tensor sizes, the performance of ByteScheduler is relatively good since partitioning large tensor sizes does not introduce dramatic extra startup overheads.

\begin{figure}[!ht]
    \captionsetup[subfigure]{justification=centering}
	\centering
	\subfloat[64 GPUs with 10GbE]{
        \includegraphics[width=0.9\columnwidth]{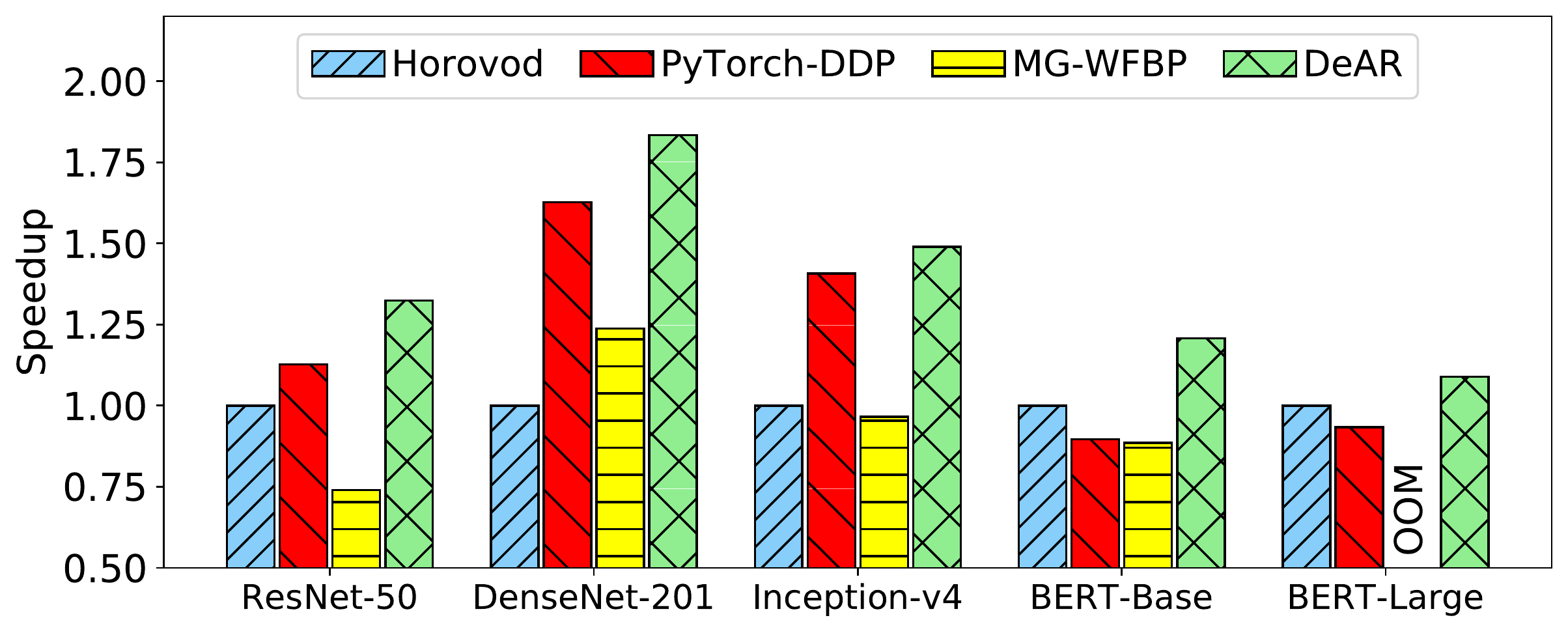}
    }\\
	\subfloat[64 GPUs with 100GbIB]{
        \includegraphics[width=0.9\columnwidth]{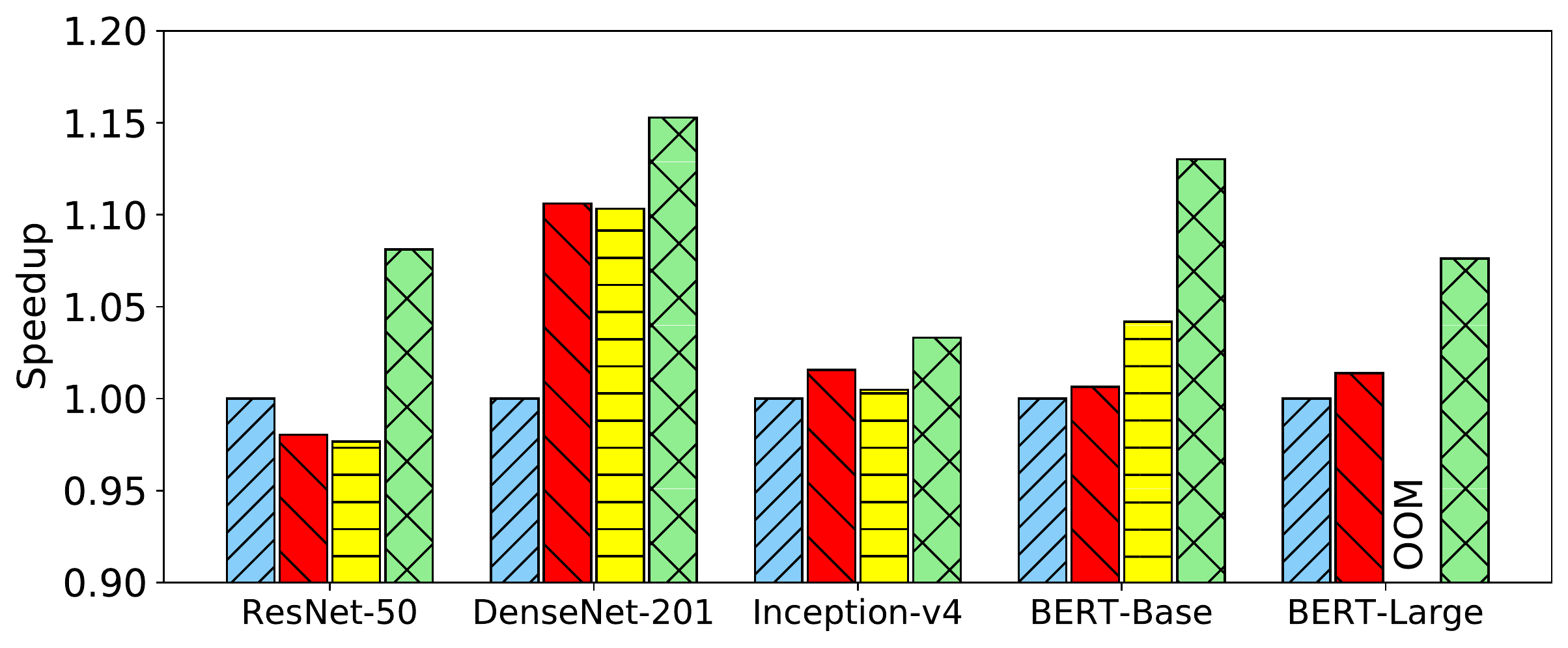}
    }
	\caption{Speedups with tensor fusion. The performance of Horovod is used as the baseline. MG-WFBP runs out-of-memory (OOM) in training BERT-Large.}
	\label{fig:speed-tf1}
\end{figure}
\subsection{Speed comparison w/ tensor fusion}
Due to the high latency of the all-to-all collectives, tensor fusion has become a default feature to accelerate distributed training. Though DeAR decouples one all-reduce operation to two operations for better communication scheduling, we need to use tensor fusion for high latency collectives (\S\ref{sec:tensorfusion}). We compare our DeAR with existing state-of-the-art algorithms including Horovod, PyTorch-DDP, and MG-WFBP~\cite{shi2019mg}, which are all equipped with tensor fusion techniques. For a fair comparison, we fix the buffer size of all algorithms with 25MB, except MG-WFBP. The results are shown in Fig.~\ref{fig:speed-tf1}, where we use the performance of Horovod as the baseline. 

\textbf{On the 10GbE cluster. }On the 64-GPU cluster with a relatively high-latency and low-bandwidth 10GbE network, we can see that DeAR with tensor fusion always outperforms all the other methods. Specifically, our DeAR achieves 6\%-83\% (an average of 36\%) improvement over existing methods on the five tested models. DeAR achieves near-linear scaling efficiency on 64 GPUs in CNNs whose number of parameters is moderate (as shown in Table~\ref{table:dnnconfig}). Though DeAR achieves significant improvement over other methods on BERT models, the scaling efficiency is still low due to the high communication-to-computation ratio. It typically requires some algorithmic-level optimizations like gradient compression~\cite{shi2021towards}. We will leave it as our future work to introduce gradient compression techniques into our DeAR scheduling framework. 

\textbf{On the 100GbIB cluster. }On the 64-GPU cluster with a low-latency and high-bandwidth 100GbIB network, the startup problem in distributed training is less significant. Even so, the end-to-end performance improvement of our DeAR over existing methods can be up to 15\% (an average of 8\%). In the 100GbIB network, the scalability of traditional methods like Horovod and PyTorch-DDP is close to linear scale in CNNs. For example, in the 64-GPU case of running ResNet-50, Horovod achieves 91\% scaling efficiency leaving limited room for further improvement. Therefore, the improvement of DeAR in 100GbIB over other methods is less significant than that of 10GbE. Given the model size and network configurations, we discuss their maximum speedups in the next subsection.

\subsection{Maximum speedups on 64-GPU clusters }\label{subsec:maxspeedups}
Intuitively, on a $P$-GPU cluster, the maximum speedup over a single GPU should be $P$, i.e., linear scale. However, due to the communication constraint, the maximum speedup may be smaller than $P$. In DeAR, each training iteration time is composed of four parts: feed-forward computation ($t_{ff}$), backpropagation computation ($t_{bp}$), gradient communication of reduce-scatter ($t_{rs}$) and gradient communication of all-gather ($t_{ag}$). The all-reduce time is $t_{ar}=t_{rs}+t_{ag}$. Given a DL model with $m$ gradient size and a cluster with $P$ workers, the communication time should be larger than the time when the link bandwidth is fully utilized. For the ring-based all-reduce algorithm, $t_{ar}\geq 2m/B$ according to Eq.~\ref{equ:time-allreduce}, where $B=1/\beta$ is the minimum link bandwidth between any two workers. Thus, for any scheduling algorithms that pipeline communications with computations, the speedup of the overall throughput on the $P$-worker system over the single worker is limited by
\begin{equation}
    S^{max}=\frac{P\times (t_{ff}+t_{bp})}{t_{ff}+t_{bp}+t_{ar}-\min\{t_{rs}, t_{bp}\} - \min\{t_{ag}, t_{ff}\}},
\end{equation}
where $\min\{t_{rs}, t_{bp}\}$ and $\min\{t_{ag}, t_{ff}\}$ are the overlapped time during backpropagation and feed-forward, respectively. Optimally, either computations or communications are fully hidden. According to the link bandwidth of 10GbE and 100GbIB and the model size shown in Table~\ref{table:dnnconfig}, we can compare the achieved speedups of DeAR on the 64-GPU cluster over a single GPU with the maximum speedups as shown in Table.~\ref{table:speedupcompare}. 
\begin{table}[!ht]
    \centering
     \caption{Comparison between the real speedup ($S$) of DeAR on 64-GPU clusters over single GPU and the theoretical maximal speedup ($S^{max}$).}
    \label{table:speedupcompare}
    \centering
    \addtolength{\tabcolsep}{-5pt}
    \begin{tabular}{|c|c|c|c|c|c|c|}
    \hline
    & & \multicolumn{5}{c|}{Model} \\\cline{1-7}
      & & ResNet-50 & DenseNet-201  & Inception-v4 & BERT-Base & BERT-Large\\\hline\hline
   \multirow{3}{*}{\shortstack[c]{10-\\GbE}} & $S^{max}$ & 61.6 & 64 & 59.8 & 25.5 & 12.1 \\\cline{2-7}
    & $S$  & 61.1 & 52.8 & 56.5 & 23.9 & 11.8 \\\cline{2-7}
    & $\frac{S}{S^{max}}$  & 99.2\% & 82.5\% & 94.5\% & 93.9\% & 98.0\% \\\hline
   \multirow{3}{*}{\shortstack[c]{100-\\GbIB}} & $S^{max}$ & 64 & 64 & 64 & 64 & 51.8 \\\cline{2-7}
    & $S$  & 61.6 & 54.0 & 57.2 & 49.6 & 37.5 \\\cline{2-7}
    & $\frac{S}{S^{max}}$  & 96.2\% & 84.4\% & 89.4\% & 77.5\% & 72.3\% \\\hline
    \end{tabular}
\end{table}

It is seen that given a high communication-to-computation ratio, the linear scaling efficiency may not be reachable. For example, on the 10GbE cluster, theoretically, running BERT models on 64 GPUs can only achieve less than 25.5 times speedup over a single GPU. In our DeAR which has two level of pipelining between communications and computations, it achieves an average of 93.6\% and 83.9\% of the theoretical optimal speedup on 10GbE and 100GbIB interconnects, respectively. 

\begin{figure}[!ht]
	\centering
	\includegraphics[width=\columnwidth]{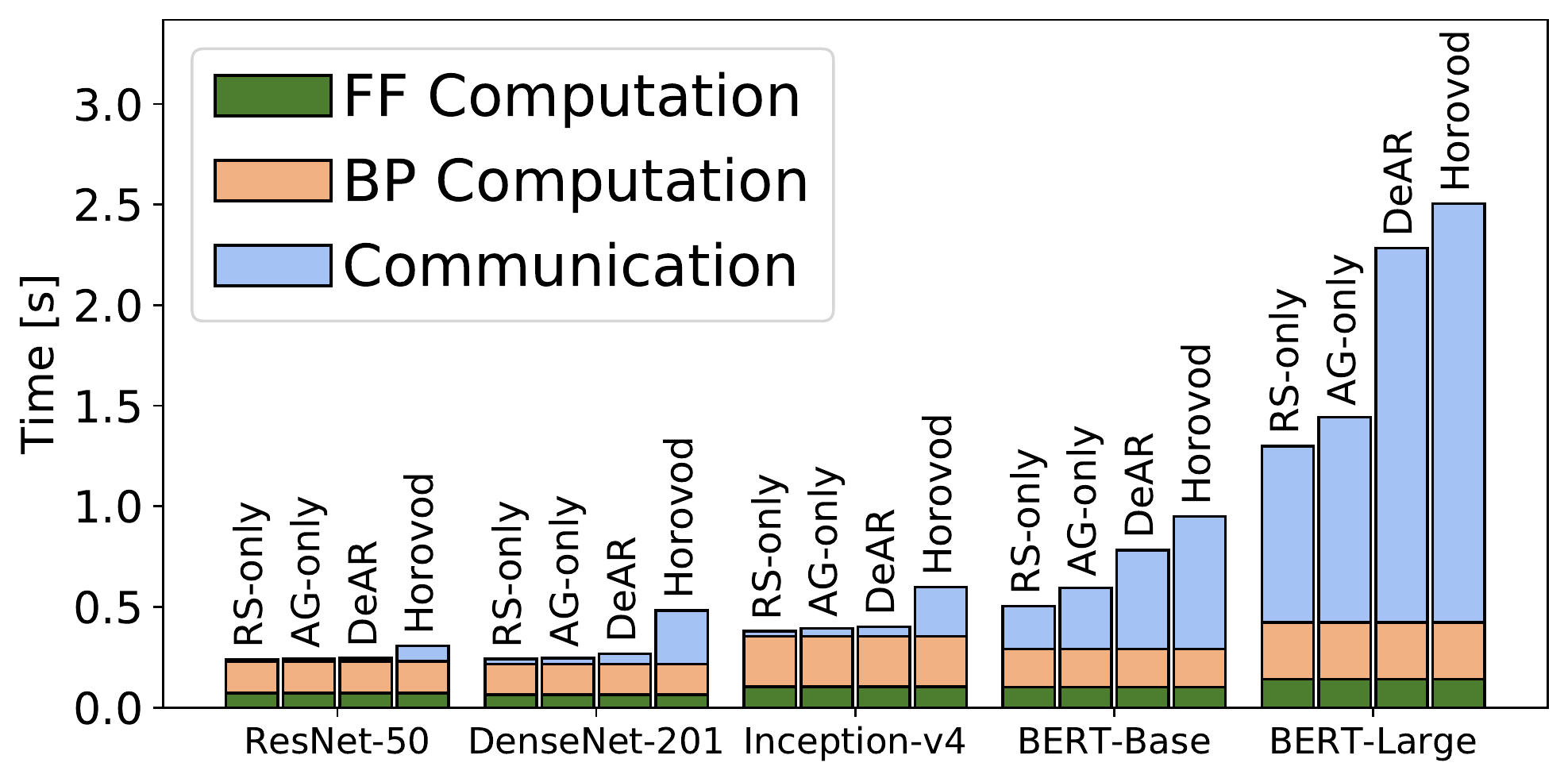}
	\caption{Time breakdowns. The communication time excludes the part hidden by computations.}
	\label{fig:time-breakdown}
\end{figure}
\subsection{Time breakdowns}
To understand the improvement of DeAR over existing methods clearer, we break down the iteration time into three parts: feed-forward (FF) computation time, backpropagation (BP) computation time, and gradient aggregation communication time (under the 10GbE network configuration) as shown in Fig.~\ref{fig:time-breakdown}. The blue bars are the non-overlapped communication time in one iteration, which means the hidden time by computations is excluded. As both DeAR and Horovod use the same back-end of PyTorch, the FF and BP computation times are the same on the same model. In DeAR, there are two parts of communication, reduce-scatter and all-gather. We use RS-only to indicate that DeAR excludes the time of all-gather, and AG-only to indicate that DeAR excludes the time of reduce-scatter in Fig.~\ref{fig:time-breakdown}.

\textbf{RS-only vs. AG-only. }In the decoupling of all-reduce, reduce-scatter and all-gather operations have the same message size for communication and they have the same communication complexity as shown in Eq.~\ref{equ:time-reducescatter} and Eq.~\ref{equ:time-allgather}. In other words, the communication time of reduce-scatter and all-gather operations should be the same without considering the overlapping with computations. However, as shown in Fig.~\ref{fig:time-breakdown}, RS-only has a less communication overhead than AG-only. The reason is that the reduce-scatter operations are overlapped with backpropagation computing tasks which typically take two times slower than feed-forward computing tasks~\cite{sze2017efficient}. Thus, reduce-scatter has more communication tasks to be overlapped with computing tasks than all-gather.

\textbf{DeAR vs. Horovod. }In the results on the breakdown of the iteration time, we can see that DeAR has a smaller non-overlapped communication time than Horovod. Though our DeAR has a similar pipelining with Horovod during backpropagation, DeAR has the opportunity to further pipeline the communications with feed-forward computations, which contributes to the performance improvement. 

\subsection{Studies of tensor fusion}
\begin{figure}[!ht]
    \captionsetup[subfigure]{justification=centering}
	\centering
	\hspace{-10pt}
	\subfloat[On the 10GbE cluster]{
        \includegraphics[width=0.485\columnwidth]{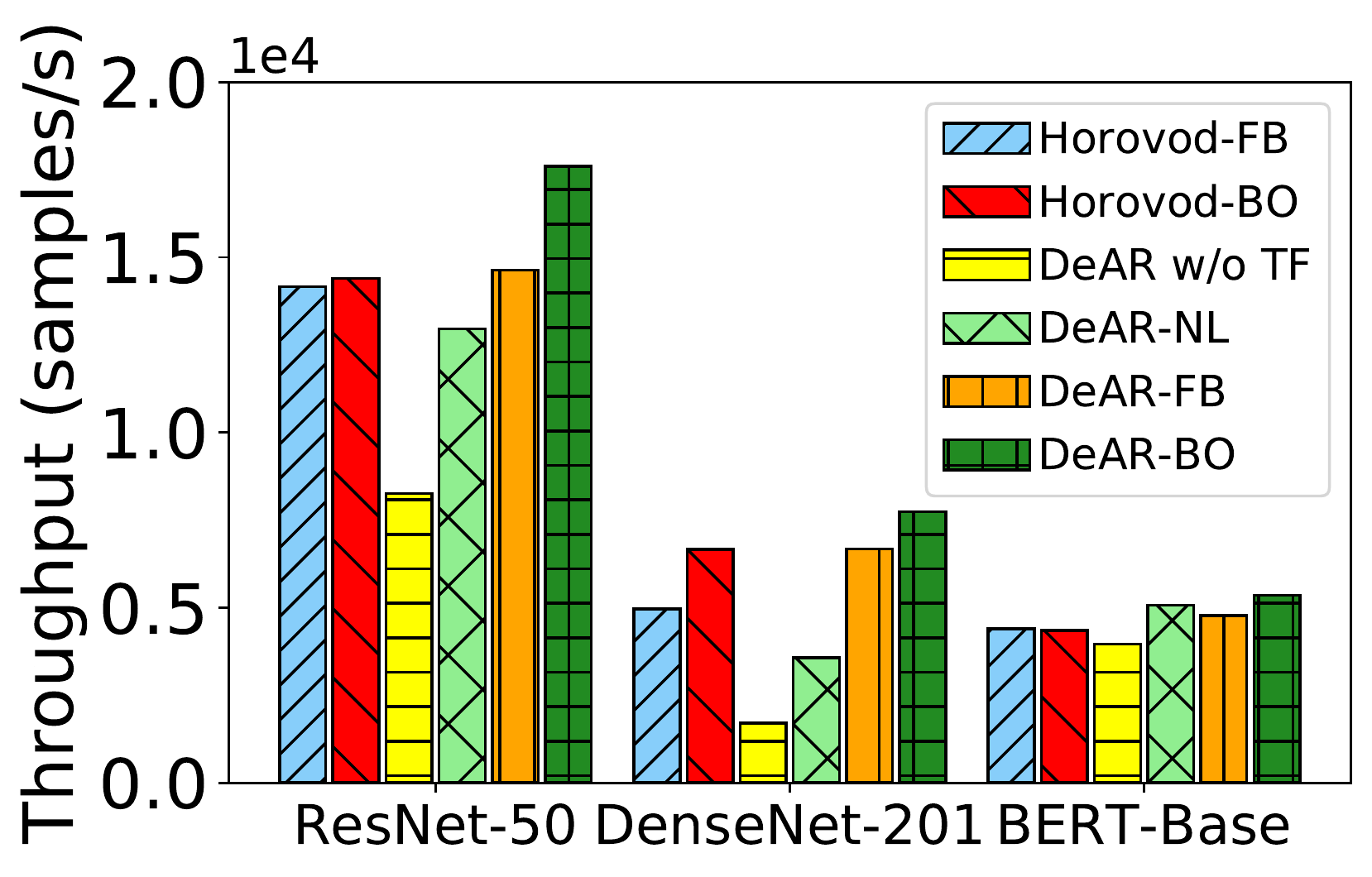}
    }
    \subfloat[On the 100GbIB cluster]{
        \includegraphics[width=0.485\columnwidth]{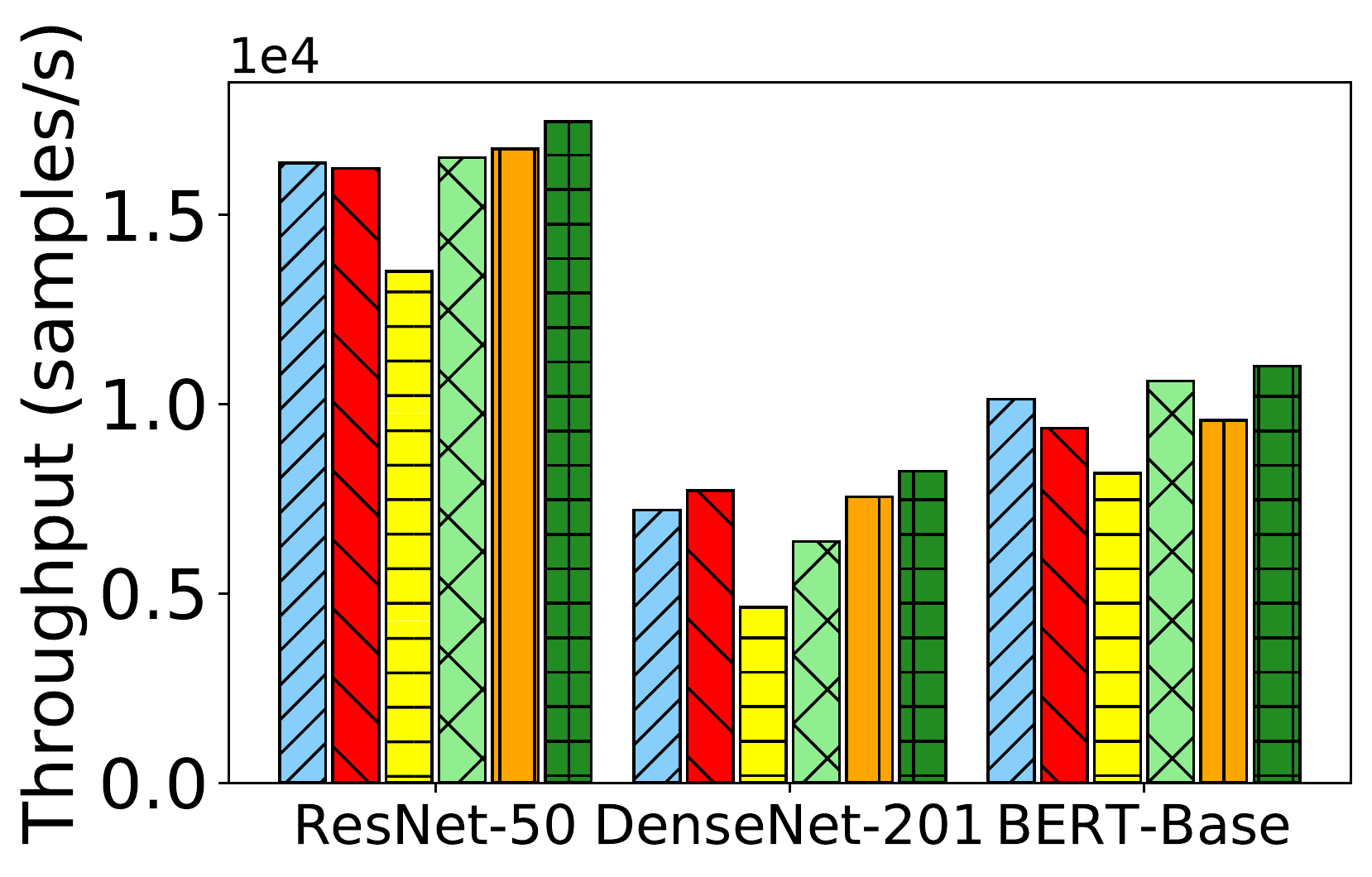}
    }
	\caption{Speed improvements with dynamic tensor fusion.}
	\label{fig:tuning-improvements}
\end{figure}

Tensor fusion is an important technique to improve performance as shown in Fig.~\ref{fig:tuning-improvements}. The tensor fusion version of DeAR with BO (DeAR-BO) achieves 1.35$\times$-4.54$\times$ and 1.29$\times$-1.78$\times$ improvement over the version w/o tensor fusion (DeAR w/o TF) under 10GbE and 100GbIB interconnects, respectively. 

\noindent \textit{1) Performance comparison between w/ BO and w/o BO}

Determining which tensors should be fused is very important in affecting the training performance. We compare our BO-based tensor fusion method (DeAR-BO) with two naive tensor fusion methods, a fixed number of four nearby layers (DeAR-NL) and a fixed buffer size (5MB) threshold (DeAR-FB), and BO-based Horovod. The results are shown in Fig.~\ref{fig:tuning-improvements}.

\textbf{Horovod-BO vs. Horovod-FB. }Horovod with BO (Horovod-BO) achieves only slight improvement in ResNet-50 and DenseNet-201 over Horovod with a fixed buffer (Horovod-FB) size (64MB by default), while Horovod-BO has no improvement in BERT-Base over Horovod-FB. The results indicate that Horovod-FB may be a good solution in tensor fusion for WFBP and tuning the buffer size does not help improve the performance. 

\textbf{DeAR-NL vs. DeAR-FB. }By merging a fixed number of consecutive layers, DeAR-NL is normally worse than Horovod-FB or DeAR-FB in CNNs as CNNs have a very imbalanced number of parameters in different layers. For the model (BERT-Base) that has a very balanced distribution of parameters in different layers, DeAR-NL performs better than Horovod-FB and DeAR-FB. 

\textbf{DeAR-FB vs. Horovod-FB. }With the opportunity of overlapping communications with both feed-forward and backpropagation computing tasks, DeAR-FB normally outperforms Horovod-FB. Since DeAR requires to pipeline tasks during feed-forward, the configured buffer size is more sensitive to the time performance than Horovod-FB. Thus, DeAR-FB only achieves a marginal improvement over Horovod-FB.

\textbf{DeAR-BO vs. others. }Using BO in DeAR, the buffer size can be well adjusted during run-time. Hence, DeAR-BO can improve the training speed over DeAR-FB, and it achieves the best performance among all the evaluated methods. Specifically, DeAR-BO is around 22\%-56\% and 7\%-14\% faster than Horovod-FB on 10GbE and 100GbIB 64-GPU clusters, respectively.

\noindent \textit{2) Search cost comparison between BO and other methods}
\begin{figure}[!t]
	\centering
	\hspace{-10pt}
	\subfloat[On the 10GbE cluster]{
        \includegraphics[width=0.48\columnwidth]{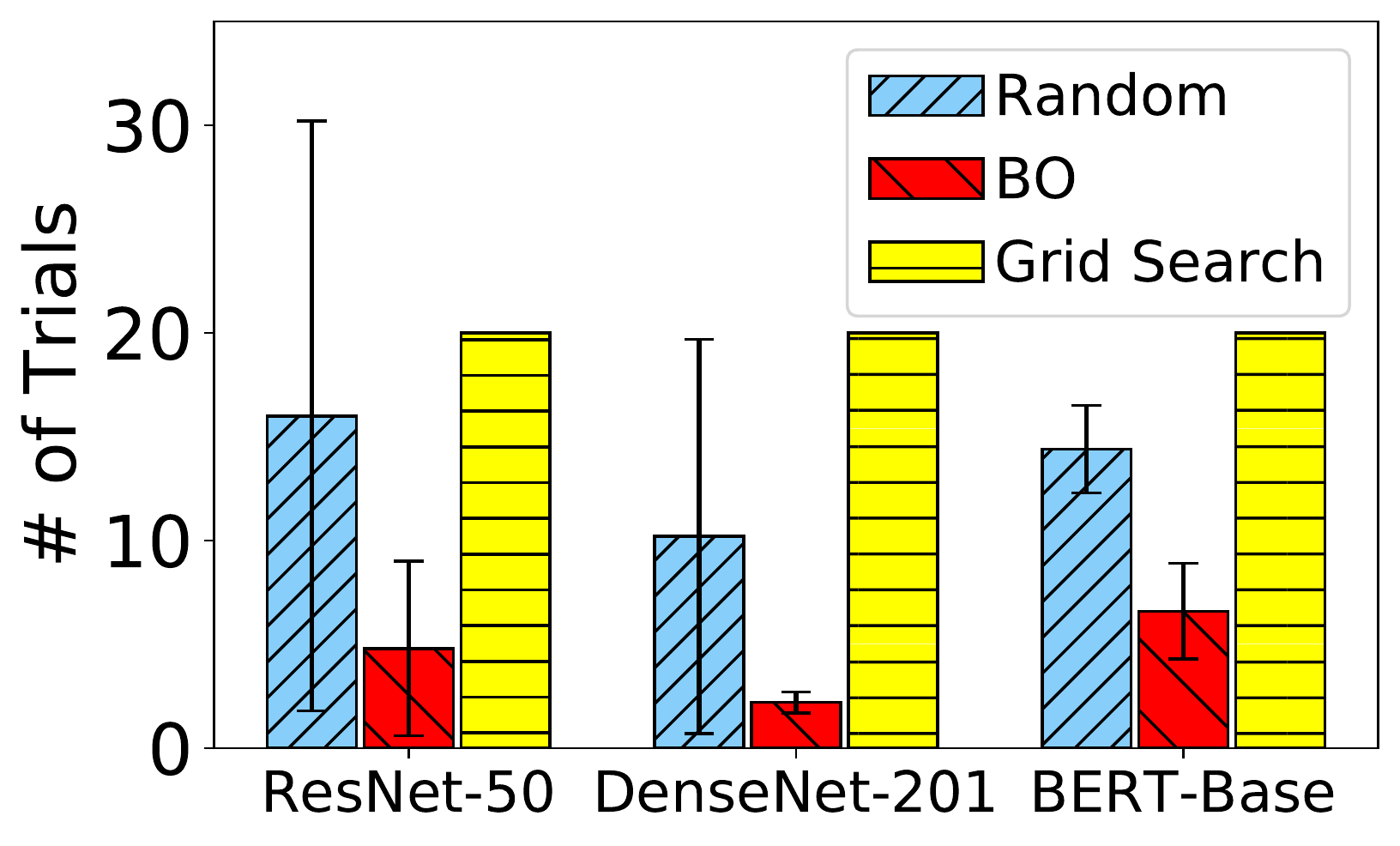}
    }
    \subfloat[On the 100GbIB cluster]{
        \includegraphics[width=0.48\columnwidth]{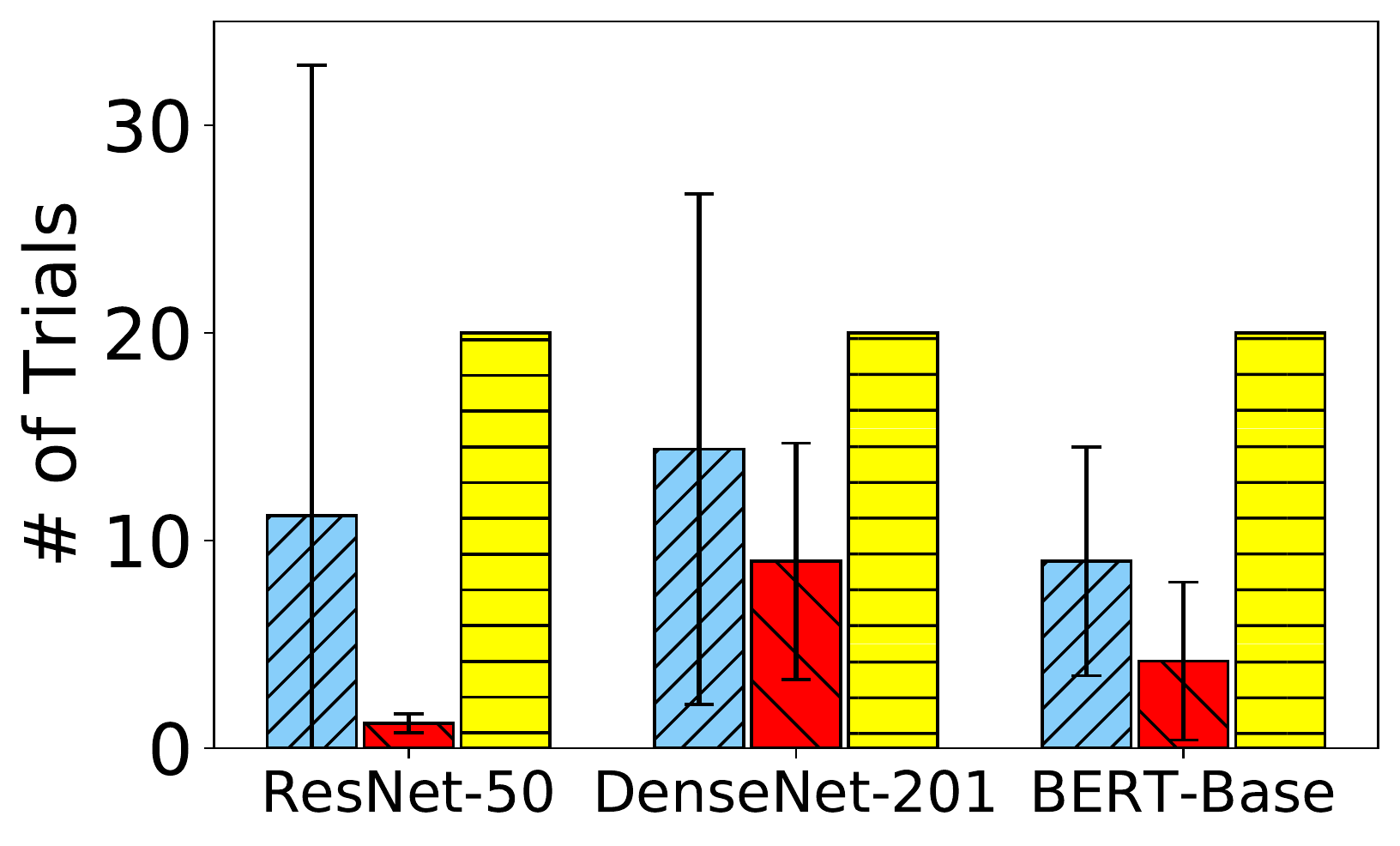}
    }
	\caption{Tuning cost of different search algorithms. Error bars show standard deviation. }
	\label{fig:tuning-cost}
\end{figure}

In DeAR, we use BO to find a good buffer size. Actually, one can also use random search or grid search to find the solution. However, both random search and grid search take a much larger number of iterations to find a good solution. For example, as shown in Fig.\ref{fig:tuning-cost}, in training different models, BO takes several trials to find a stable solution while random and grid search take tens of trials. In DNN training, one cannot consume too much number of iterations to tune their time performance related parameters, otherwise it may result in a longer end-to-end training time than the naive methods. The average cost of BO is 0.207 seconds per trial over 20 trials.

\subsection{Performance with different batch sizes}
\begin{figure}[!t]
	\centering
	\subfloat[ResNet-50]{
        \includegraphics[width=0.48\columnwidth]{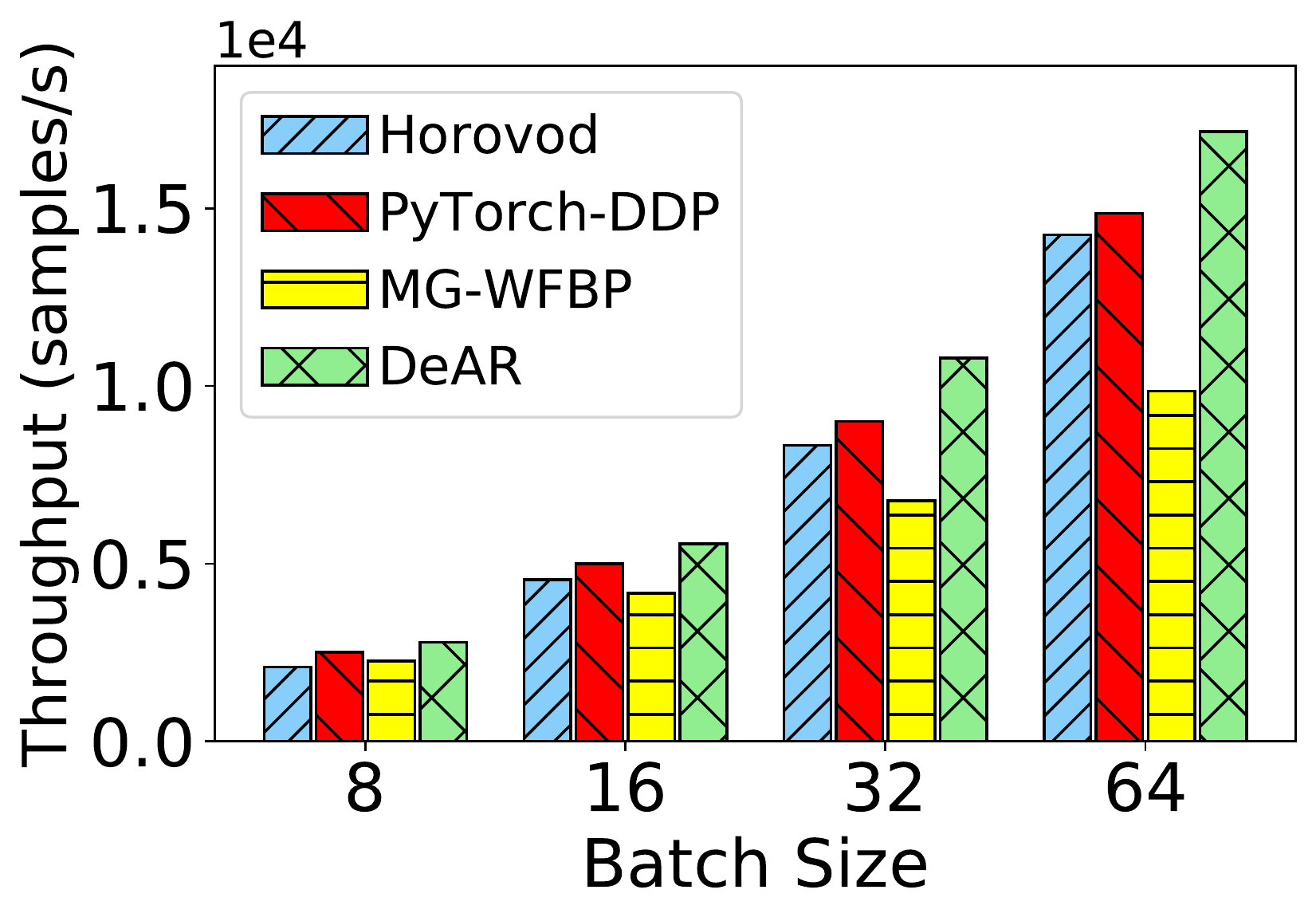}
    }
    \subfloat[BERT-Base]{
        \includegraphics[width=0.46\columnwidth]{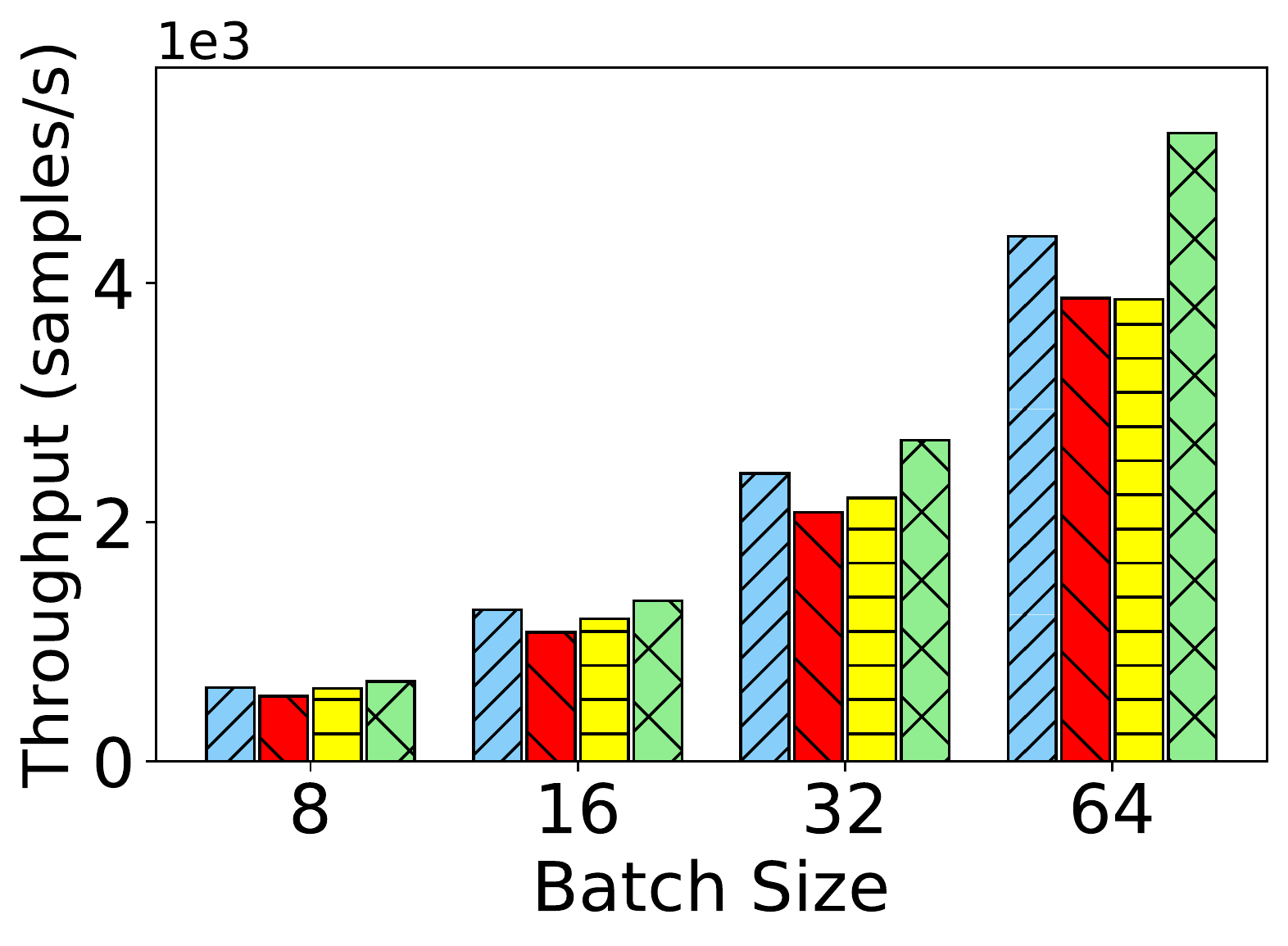}
    }
	\caption{Speed comparison with different batch sizes.}
	\label{fig:speed-bs}
\end{figure}

The local mini-batch size has a direct impact on the feed-forward and backpropagation computation time, thus training with different batch sizes has different communication-to-computation ratios on the same model, which would affect the opportunity for overlapping. We compare our DeAR with the existing methods on the 10GbE cluster using ResNet-50 and BERT-Base under different batch sizes as shown in Fig.~\ref{fig:speed-bs}. The batch size is the local mini-batch size for each GPU running on the 64-GPU clusters. Smaller batch size indicates shorter computation time while the communication time remains unchanged (if not overlapped). The results show that our DeAR is robust to the mini-batch size and it outperforms all other methods in all tested cases.

\subsection{Potential improvement on larger-scale clusters}
Due to the hardware limit, we are unable to perform experiments on larger-scale GPU clusters to verify the performance of DeAR. We provide some discussions here. According to Fig.~\ref{fig:speed-tf1}, we can see that the improvement of our DeAR over existing methods on 100GbIB (an average of 8\%) is much smaller than on 10GbE (an average of 36\%). In our experiment environments, the two clusters are with the same GPU hardware, which means the FF and BP computation time remain unchanged while the network speed of 100GbIB is 10 times faster than 10GbE. That is, existing methods should be better scaling efficiency with a higher network speed, which makes the optimization room be smaller. For example, in ResNet-50, Horovod achieves 58.2$\times$ speedup on 64 GPUs with 100GbIB over the single GPU while the maximum speedup is only 64$\times$ (Table~\ref{table:speedupcompare}). Therefore, any optimization cannot achieve improvement larger than 10\% over Horovod. We argue that with increasing size of the cluster and more powerful GPUs, DeAR could have a higher improvement over Horovod and PyTorch-DDP as the communication-to-computation ratio becomes larger. 

\change{Formally, the theoretical optimal iteration time for DeAR and baseline algorithms with perfect overlapping is given by}
\begin{align}
    t_{DeAR} = \max\{t_{ff}, t_{ag}\} + \max\{t_{bp}, t_{rs}\}, \\
    t_{baseline} = t_{ff} + \max\{t_{bp}, t_{ar}\}, 
\end{align}
\change{where $t_{ff}$ and $t_{bp}$ are feed-forward and back-propagation computation time, $t_{rs}$, $t_{ag}$, and $t_{ar}$ are reduce-scatter, all-gather, and all-reduce communication time, respectively. Assume that $t_{ar}=2t_{rs}=2t_{ag}$ and $t_{bp}=2t_{ff}$, we can derive that}
\begin{align}
    t_{baseline} - t_{DeAR} = 
    \begin{cases}
    0, &\text{if}~t_{ag} \le t_{ff}, \\
    t_{ag}-t_{ff}, &\text{if}~t_{ff} < t_{ag} \le 2t_{ff}, \\ 
    t_{ff}, &\text{otherwise.}
    \end{cases}
\end{align}
\change{This implies that, if we implement DeAR properly to overlap communications with forward and backward computations, DeAR can always outperform baseline algorithms such as Horovod and PyTorch-DDP. As the communication-to-computation ratio becomes larger, the saved iteration time can be at most one feed-forward computation cost of $t_{ff}$. } 

\section{Related Work}\label{sec:relatedwork}
There are many studies in addressing the communication problem of distributed training, like asynchronous training~\cite{zheng2017asynchronous,lian2015asynchronous} and gradient compression~\cite{alistarh2017qsgd,lin2018deep,shi2019convergence,basu2019qsparse}. A more comprehensive introduction can be found in recent survey papers~\cite{tang2020communication,ben2019demystifying}. Here we highlight some very related work from the systems perspective in S-SGD. 

\subsection{Efficient all-reduce design}
As the all-reduce collective is very ubiquitous in distributed deep learning, there are many works providing efficient algorithms for different cluster configurations\cite{sanders2009two,awan2016efficient,awan2017s,jia2018highly,cho2019blueconnect,luo2020plink,wang2020blink,dong2020eflops,shi2021towards,miao2021heterogeneity}. For example, the double-binary trees algorithm~\cite{sanders2009two} has been integrated in NCCL to scale-out on extremely large-scale GPU clusters. Some particular all-reduce algorithms are also designed for different cluster topology, like~\cite{mikami2018massively} on Torus networks, \cite{cho2019blueconnect,wang2020blink} on NVLink-based GPU servers, \cite{luo2020plink,shi2021towards} on public cloud clusters, and \cite{miao2021heterogeneity} on heterogeneous GPU clusters. These designs are orthogonal to our DeAR as long as these algorithms can be decoupled into two operations without introducing extra overheads. \change{For example, one can decompose the double-binary tree-based all-reduce~\cite{sanders2009two} into tree-based reduce and tree-based broadcast, and decompose the hierarchical ring-based all-reduce~\cite{mikami2018massively} into intra-node and inter-node reduce-scatter and all-gather. We leave decoupling more all-reduce algorithms as our future work, and the decoupling configuration can be automatically tuned using BO. }

\subsection{Communication scheduling}
Due to the layer-wise structure of DNN models and even tensor-wise in modern deep learning frameworks like PyTorch and TensorFlow, the computing and communication tasks are generally organized as a directed acyclic graph (DAG). Thus, the tasks without any dependency can be executed concurrently, making it possible to hide some communication costs by overlapping them with computing tasks. The wait-free backpropagation (WFPB) algorithm~\cite{zhang2017poseidon,awan2017s} was the early scheduling method that pipelines the communication tasks of gradient aggregation with gradient calculation during backpropagation. Then some tensor fusion techniques (e.g.,~\cite{you2017scaling,shi2019mg,shi2021mg,shi2021exploiting,romero2022accelerating}) were further proposed to address the high latency problem in WFBP with all-reduce. As the communication of some large tensors may block the execution of higher-priority tensors, ByteScheduler~\cite{peng2019generic} proposes tensor partitioning in priority scheduling to provide a finer-grained schedule of overlapping between computing and communication tasks. ZeRO~\cite{rajbhandari2020zero} decoupled all-reduce like DeAR, but it was to shard parameters to optimize memory rather than optimizing communication efficiency. \change{To shard parameters, ZeRO requires one all-gather for each forward pass and one extra all-gather for each backward pass, which unfortunately has increased the total communication overheads compared with DeAR. In the recent PyTorch v1.13 release, FullyShardedDataParallel\footnote{\url{https://pytorch.org/docs/stable/fsdp.html}} has combined the ideas of ZeRO's parameter sharding and DeAR's FeedPipe to alleviate the memory and communication overheads, respectively, but it does not consider dynamic tensor fusion to explore the optimal training performance. }

\section{Conclusion}\label{sec:conclusion}
In this work, we proposed a novel communication scheduling algorithm named DeAR with fine-grained all-reduce pipelining. In DeAR, we first decoupled the all-reduce primitive into two continuous communication operations without introducing any extra communication overhead to enable a fine-grained schedule of computations and communications. Then we proposed to pipeline the backpropagation and feed-forward computing tasks with the first operations and second operations of the decoupled all-reduce primitives, respectively. To integrate the effective tensor fusion technique, we proposed a practical tensor fusion method using Bayesian optimization in DeAR to further reduce the communication time without considering the DNN model and network configuration. Extensive experiments were conducted on a 64-GPU cluster connected with two types of networks (10Gb/s Ethernet and 100Gb/s InfiniBand) on different applications including CNNs and BERTs. Experimental results show that DeAR achieves up to 83\% improvement over existing state-of-the-art methods.

\section*{Acknowledgments}
The research was supported in part by a RGC RIF grant under the contract R6021-20, RGC GRF grants under the contracts 16209120, 16200221, 16207922, 16213120, the National Natural Science Foundation of China (NSFC) (Grant No. 62272122), and the National Natural Science Foundation of China (NSFC) (Grant No. 62002240).
\bibliographystyle{IEEEtran}
\bibliography{cites}

\begin{thebibliography}{10}
\providecommand{\url}[1]{#1}
\csname url@samestyle\endcsname
\providecommand{\newblock}{\relax}
\providecommand{\bibinfo}[2]{#2}
\providecommand{\BIBentrySTDinterwordspacing}{\spaceskip=0pt\relax}
\providecommand{\BIBentryALTinterwordstretchfactor}{4}
\providecommand{\BIBentryALTinterwordspacing}{\spaceskip=\fontdimen2\font plus
\BIBentryALTinterwordstretchfactor\fontdimen3\font minus
  \fontdimen4\font\relax}
\providecommand{\BIBforeignlanguage}[2]{{%
\expandafter\ifx\csname l@#1\endcsname\relax
\typeout{** WARNING: IEEEtran.bst: No hyphenation pattern has been}%
\typeout{** loaded for the language `#1'. Using the pattern for}%
\typeout{** the default language instead.}%
\else
\language=\csname l@#1\endcsname
\fi
#2}}
\providecommand{\BIBdecl}{\relax}
\BIBdecl

\bibitem{dean2012large}
J.~Dean, G.~Corrado, R.~Monga, K.~Chen, M.~Devin, M.~Mao, M.~Ranzato,
  A.~Senior, P.~Tucker, K.~Yang \emph{et~al.}, ``Large scale distributed deep
  networks,'' in \emph{Proc. of NeurIPS}, 2012, pp. 1223--1231.

\bibitem{jia2018highly}
X.~Jia, S.~Song, S.~Shi, W.~He, Y.~Wang, H.~Rong, F.~Zhou, L.~Xie, Z.~Guo,
  Y.~Yang, L.~Yu, T.~Chen, G.~Hu, and X.~Chu, ``Highly scalable deep learning
  training system with mixed-precision: Training {ImageNet} in four minutes,''
  in \emph{Proc. of Workshop on Systems for ML and Open Source Software,
  collocated with NeurIPS 2018}, 2018.

\bibitem{you2020large}
Y.~You, J.~Li, S.~Reddi, J.~Hseu, S.~Kumar, S.~Bhojanapalli, X.~Song,
  J.~Demmel, K.~Keutzer, and C.-J. Hsieh, ``Large batch optimization for deep
  learning: Training bert in 76 minutes,'' in \emph{International Conference on
  Learning Representations}, 2020.

\bibitem{goyal2017accurate}
P.~Goyal, P.~Doll{\'a}r, R.~Girshick, P.~Noordhuis, L.~Wesolowski, A.~Kyrola,
  A.~Tulloch, Y.~Jia, and K.~He, ``Accurate, large minibatch {SGD}: Training
  {ImageNet} in 1 hour,'' \emph{arXiv preprint arXiv:1706.02677}, 2017.

\bibitem{huang2019gpipe}
Y.~Huang, Y.~Cheng, A.~Bapna, O.~Firat, D.~Chen, M.~Chen, H.~Lee, J.~Ngiam,
  Q.~V. Le, Y.~Wu \emph{et~al.}, ``{GPipe}: Efficient training of giant neural
  networks using pipeline parallelism,'' \emph{Proc. of NeurIPS}, vol.~32,
  2019.

\bibitem{narayanan2021efficient}
D.~Narayanan, M.~Shoeybi, J.~Casper, P.~LeGresley, M.~Patwary, V.~Korthikanti,
  D.~Vainbrand, P.~Kashinkunti, J.~Bernauer, B.~Catanzaro \emph{et~al.},
  ``Efficient large-scale language model training on {GPU} clusters using
  {megatron-LM},'' in \emph{Proc. of SC}, 2021, pp. 1--15.

\bibitem{awan2016efficient}
A.~A. Awan, K.~Hamidouche, A.~Venkatesh, and D.~K. Panda, ``Efficient large
  message broadcast using {NCCL} and {CUDA-aware MPI} for deep learning,'' in
  \emph{Proc. of The 23rd European MPI Users' Group Meeting}, 2016, pp. 15--22.

\bibitem{cho2019blueconnect}
M.~Cho, U.~Finkler, and D.~Kung, ``Blueconnect: Novel hierarchical all-reduce
  on multi-tired network for deep learning,'' in \emph{Proceedings of the
  Conference on Systems and Machine Learning (SysML)}, 2019.

\bibitem{chu2020nv}
C.-H. Chu, P.~Kousha, A.~A. Awan, K.~S. Khorassani, H.~Subramoni, and D.~K.
  Panda, ``{NV-group}: link-efficient reduction for distributed deep learning
  on modern dense {GPU} systems,'' in \emph{Proceedings of the 34th ACM
  International Conference on Supercomputing}, 2020, pp. 1--12.

\bibitem{shi2020quantitative}
S.~Shi, Z.~Tang, X.~Chu, C.~Liu, W.~Wang, and B.~Li, ``A quantitative survey of
  communication optimizations in distributed deep learning,'' \emph{IEEE
  Network}, vol.~35, no.~3, pp. 230--237, 2020.

\bibitem{xu2017performance}
P.~Xu, S.~Shi, and X.~Chu, ``Performance evaluation of deep learning tools in
  docker containers,'' in \emph{2017 3rd International Conference on Big Data
  Computing and Communications (BIGCOM)}.\hskip 1em plus 0.5em minus
  0.4em\relax IEEE, 2017, pp. 395--403.

\bibitem{shi2018performance}
S.~Shi, W.~Qiang, and X.~Chu, ``Performance modeling and evaluation of
  distributed deep learning frameworks on {GPUs},'' in \emph{Proc. of The 4th
  International Conference on Big Data Intelligence and Computing}.\hskip 1em
  plus 0.5em minus 0.4em\relax IEEE, 2018.

\bibitem{zhang2017poseidon}
H.~Zhang, Z.~Zheng, S.~Xu, W.~Dai, Q.~Ho, X.~Liang, Z.~Hu, J.~Wei, P.~Xie, and
  E.~P. Xing, ``Poseidon: An efficient communication architecture for
  distributed deep learning on {GPU} clusters,'' in \emph{2017 USENIX Annual
  Technical Conference (USENIX ATC 17)}, 2017, pp. 181--193.

\bibitem{awan2017s}
A.~A. Awan, K.~Hamidouche, J.~M. Hashmi, and D.~K. Panda, ``S-caffe:
  Co-designing {MPI} runtimes and {Caffe} for scalable deep learning on modern
  {GPU} clusters,'' in \emph{Proceedings of the 22nd ACM SIGPLAN Symposium on
  Principles and Practice of Parallel Programming}, 2017, pp. 193--205.

\bibitem{pytorchddp}
S.~Li, Y.~Zhao, R.~Varma \emph{et~al.}, ``Pytorch distributed: Experiences on
  accelerating data parallel training,'' \emph{Proc. of VLDB}, vol.~13, no.~12.

\bibitem{sergeev2018horovod}
A.~Sergeev and M.~D. Balso, ``Horovod: fast and easy distributed deep learning
  in {TensorFlow},'' \emph{arXiv preprint arXiv:1802.05799}, 2018.

\bibitem{romero2022accelerating}
J.~Romero, J.~Yin, N.~Laanait, B.~Xie, M.~T. Young, S.~Treichler,
  V.~Starchenko, A.~Borisevich, A.~Sergeev, and M.~Matheson, ``Accelerating
  collective communication in data parallel training across deep learning
  frameworks,'' in \emph{Proc. of NSDI}, 2022, pp. 1027--1040.

\bibitem{sze2017efficient}
V.~Sze, Y.-H. Chen, T.-J. Yang, and J.~S. Emer, ``Efficient processing of deep
  neural networks: A tutorial and survey,'' \emph{Proceedings of the IEEE},
  vol. 105, no.~12, pp. 2295--2329, 2017.

\bibitem{barnett1994interprocessor}
M.~Barnett, L.~Shuler, R.~van~de Geijn, S.~Gupta, D.~Payne, and J.~Watts,
  ``Interprocessor collective communication library (intercom),'' in
  \emph{Proceedings of IEEE Scalable High Performance Computing Conference},
  1994, pp. 357--364.

\bibitem{rabenseifner2004optimization}
R.~Rabenseifner, ``Optimization of collective reduction operations,'' in
  \emph{International Conference on Computational Science}.\hskip 1em plus
  0.5em minus 0.4em\relax Springer, 2004, pp. 1--9.

\bibitem{thakur2005optimization}
R.~Thakur, R.~Rabenseifner, and W.~Gropp, ``Optimization of collective
  communication operations in {MPICH},'' \emph{The International Journal of
  High Performance Computing Applications}, vol.~19, no.~1, pp. 49--66, 2005.

\bibitem{hoefler2010toward}
T.~Hoefler, W.~Gropp, R.~Thakur, and J.~L. Tr{\"a}ff, ``Toward performance
  models of {MPI} implementations for understanding application scaling
  issues,'' in \emph{European MPI Users' Group Meeting}.\hskip 1em plus 0.5em
  minus 0.4em\relax Springer, 2010, pp. 21--30.

\bibitem{shi2019mg}
S.~Shi, X.~Chu, and B.~Li, ``{MG-WFBP}: Efficient data communication for
  distributed synchronous {SGD} algorithms,'' in \emph{IEEE INFOCOM 2019-IEEE
  Conference on Computer Communications}.\hskip 1em plus 0.5em minus
  0.4em\relax IEEE, 2019, pp. 172--180.

\bibitem{shi2021mg}
------, ``{MG-WFBP}: Merging gradients wisely for efficient communication in
  distributed deep learning,'' \emph{IEEE Transactions on Parallel and
  Distributed Systems}, vol.~32, no.~8, pp. 1903--1917, 2021.

\bibitem{peng2019generic}
Y.~Peng, Y.~Zhu, Y.~Chen, Y.~Bao, B.~Yi, C.~Lan, C.~Wu, and C.~Guo, ``A generic
  communication scheduler for distributed {DNN} training acceleration,'' in
  \emph{Proc. of SOSP}, 2019, pp. 16--29.

\bibitem{li2014scaling}
M.~Li, D.~G. Andersen, J.~W. Park, A.~J. Smola, A.~Ahmed, V.~Josifovski,
  J.~Long, E.~J. Shekita, and B.-Y. Su, ``Scaling distributed machine learning
  with the parameter server.'' in \emph{OSDI}, vol.~14, 2014, pp. 583--598.

\bibitem{patarasuk2009bandwidth}
P.~Patarasuk and X.~Yuan, ``Bandwidth optimal all-reduce algorithms for
  clusters of workstations,'' \emph{Journal of Parallel and Distributed
  Computing}, vol.~69, no.~2, pp. 117--124, 2009.

\bibitem{thakur2003improving}
R.~Thakur and W.~D. Gropp, ``Improving the performance of collective operations
  in {MPICH},'' in \emph{European Parallel Virtual Machine/Message Passing
  Interface Users’ Group Meeting}.\hskip 1em plus 0.5em minus 0.4em\relax
  Springer, 2003, pp. 257--267.

\bibitem{snoek2012practical}
J.~Snoek, H.~Larochelle, and R.~P. Adams, ``Practical bayesian optimization of
  machine learning algorithms,'' \emph{Proc. of NeurIPS}, vol.~25, 2012.

\bibitem{package2014bo}
\BIBentryALTinterwordspacing
F.~Nogueira, ``{Bayesian Optimization}: Open source constrained global
  optimization tool for {Python},'' 2014--. [Online]. Available:
  \url{https://github.com/fmfn/BayesianOptimization}
\BIBentrySTDinterwordspacing

\bibitem{Alipourfard2017CherryPick}
O.~Alipourfard, H.~H. Liu, J.~Chen, S.~Venkataraman, M.~Yu, and M.~Zhang,
  ``Cherrypick: Adaptively unearthing the best cloud configurations for big
  data analytics,'' in \emph{NSDI}, 2017.

\bibitem{huang2017densely}
G.~Huang, Z.~Liu, L.~Van Der~Maaten, and K.~Q. Weinberger, ``Densely connected
  convolutional networks,'' in \emph{Proceedings of the IEEE conference on
  computer vision and pattern recognition}, 2017, pp. 4700--4708.

\bibitem{deng2009imagenet}
J.~Deng, W.~Dong, R.~Socher, L.-J. Li, K.~Li, and L.~Fei-Fei, ``{ImageNet}: A
  large-scale hierarchical image database,'' in \emph{Proc. of CVPR}, 2009, pp.
  248--255.

\bibitem{devlin2019bert}
J.~Devlin, M.-W. Chang, K.~Lee, and K.~Toutanova, ``{BERT}: Pre-training of
  deep bidirectional transformers for language understanding,'' in
  \emph{Proceedings of the 2019 Conference of the North American Chapter of the
  Association for Computational Linguistics: Human Language Technologies,
  Volume 1 (Long and Short Papers)}, 2019, pp. 4171--4186.

\bibitem{he2016deep}
K.~He, X.~Zhang, S.~Ren, and J.~Sun, ``Deep residual learning for image
  recognition,'' in \emph{Proc. of CVPR}, 2016, pp. 770--778.

\bibitem{szegedy2017inception}
C.~Szegedy, S.~Ioffe, V.~Vanhoucke, and A.~A. Alemi, ``Inception-v4,
  inception-resnet and the impact of residual connections on learning,'' in
  \emph{Proc. of The 31st AAAI}, 2017.

\bibitem{shi2021towards}
S.~Shi, X.~Zhou, S.~Song, X.~Wang, Z.~Zhu, X.~Huang, X.~Jiang, F.~Zhou, Z.~Guo,
  L.~Xie \emph{et~al.}, ``Towards scalable distributed training of deep
  learning on public cloud clusters,'' \emph{Proceedings of Machine Learning
  and Systems}, vol.~3, pp. 401--412, 2021.

\bibitem{zheng2017asynchronous}
S.~Zheng, Q.~Meng, T.~Wang, W.~Chen, N.~Yu, Z.-M. Ma, and T.-Y. Liu,
  ``Asynchronous stochastic gradient descent with delay compensation,'' in
  \emph{International Conference on Machine Learning}, 2017, pp. 4120--4129.

\bibitem{lian2015asynchronous}
X.~Lian, Y.~Huang, Y.~Li, and J.~Liu, ``Asynchronous parallel stochastic
  gradient for nonconvex optimization,'' in \emph{Proc. of NeurIPS}, 2015, pp.
  2737--2745.

\bibitem{alistarh2017qsgd}
D.~Alistarh, D.~Grubic, J.~Li, R.~Tomioka, and M.~Vojnovic, ``{QSGD}:
  Communication-efficient {SGD} via gradient quantization and encoding,'' in
  \emph{Proc. of NeurIPS}, 2017, pp. 1709--1720.

\bibitem{lin2018deep}
Y.~Lin, S.~Han, H.~Mao, Y.~Wang, and W.~J. Dally, ``Deep gradient compression:
  Reducing the communication bandwidth for distributed training,'' in
  \emph{International Conference on Learning Representations}, 2018.

\bibitem{shi2019convergence}
S.~Shi, K.~Zhao, Q.~Wang, Z.~Tang, and X.~Chu, ``A convergence analysis of
  distributed {SGD} with communication-efficient gradient sparsification.'' in
  \emph{IJCAI}, 2019, pp. 3411--3417.

\bibitem{basu2019qsparse}
D.~Basu, D.~Data, C.~Karakus, and S.~Diggavi, ``{Qsparse-local-SGD}:
  Distributed {SGD} with quantization, sparsification and local computations,''
  in \emph{Proc. of NeurIPS}, 2019, pp. 14\,695--14\,706.

\bibitem{tang2020communication}
Z.~Tang, S.~Shi, X.~Chu, W.~Wang, and B.~Li, ``Communication-efficient
  distributed deep learning: A comprehensive survey,'' \emph{arXiv preprint
  arXiv:2003.06307}, 2020.

\bibitem{ben2019demystifying}
T.~Ben-Nun and T.~Hoefler, ``Demystifying parallel and distributed deep
  learning: An in-depth concurrency analysis,'' \emph{ACM Computing Surveys
  (CSUR)}, vol.~52, no.~4, pp. 1--43, 2019.

\bibitem{sanders2009two}
P.~Sanders, J.~Speck, and J.~L. Tr{\"a}ff, ``Two-tree algorithms for full
  bandwidth broadcast, reduction and scan,'' \emph{Parallel Computing},
  vol.~35, no.~12, pp. 581--594, 2009.

\bibitem{luo2020plink}
L.~Luo, P.~West, J.~Nelson, A.~Krishnamurthy, and L.~Ceze, ``{PLink}:
  Discovering and exploiting locality for accelerated distributed training on
  the public cloud,'' in \emph{Proceedings of Machine Learning and Systems
  2020}, 2020, pp. 82--97.

\bibitem{wang2020blink}
G.~Wang, S.~Venkataraman, A.~Phanishayee, N.~Devanur, J.~Thelin, and I.~Stoica,
  ``Blink: Fast and generic collectives for distributed {ML},'' in
  \emph{Proceedings of Machine Learning and Systems 2020}, 2020, pp. 172--186.

\bibitem{dong2020eflops}
J.~Dong, Z.~Cao, T.~Zhang, J.~Ye, S.~Wang, F.~Feng, L.~Zhao, X.~Liu, L.~Song,
  L.~Peng \emph{et~al.}, ``{EFLOPS}: Algorithm and system co-design for a high
  performance distributed training platform,'' in \emph{2020 IEEE International
  Symposium on High Performance Computer Architecture (HPCA)}.\hskip 1em plus
  0.5em minus 0.4em\relax IEEE, 2020, pp. 610--622.

\bibitem{miao2021heterogeneity}
X.~Miao, X.~Nie, Y.~Shao, Z.~Yang, J.~Jiang, L.~Ma, and B.~Cui,
  ``Heterogeneity-aware distributed machine learning training via partial
  reduce,'' in \emph{Proceedings of the 2021 International Conference on
  Management of Data}, 2021, pp. 2262--2270.

\bibitem{mikami2018massively}
H.~Mikami, H.~Suganuma, Y.~Tanaka, Y.~Kageyama \emph{et~al.}, ``Massively
  distributed sgd: Imagenet/resnet-50 training in a flash,'' \emph{arXiv
  preprint arXiv:1811.05233}, 2018.

\bibitem{you2017scaling}
Y.~You, A.~Bulu{\c{c}}, and J.~Demmel, ``Scaling deep learning on {GPU} and
  knights landing clusters,'' in \emph{Proc. of SC}, 2017, pp. 1--12.

\bibitem{shi2021exploiting}
S.~Shi, X.~Chu, and B.~Li, ``Exploiting simultaneous communications to
  accelerate data parallel distributed deep learning,'' in \emph{IEEE INFOCOM
  2021-IEEE Conference on Computer Communications}.\hskip 1em plus 0.5em minus
  0.4em\relax IEEE, 2021, pp. 1--10.

\bibitem{rajbhandari2020zero}
S.~Rajbhandari, J.~Rasley, O.~Ruwase, and Y.~He, ``Zero: Memory optimizations
  toward training trillion parameter models,'' in \emph{Proc. of SC}.\hskip 1em
  plus 0.5em minus 0.4em\relax IEEE, 2020, pp. 1--16.

\end{thebibliography}

\end{document}